\documentclass{article} 
\usepackage{iclr2024_conference,times}
\usepackage{algorithm}
\usepackage{algorithmic}

\usepackage{enumitem}
\usepackage{longtable}

\usepackage{amsmath,amsthm,amsfonts,amssymb}
\usepackage{bm}
\usepackage{graphicx}
\usepackage{sidecap}
\usepackage{mathrsfs}
\usepackage{multirow, multicol}
\usepackage{caption}
\usepackage{wrapfig}
\usepackage{booktabs} 
\usepackage{xcolor}
\usepackage{subcaption}

\def\eqref#1{equation~\ref{#1}}

\def\1{\bm{1}}

\def\vtheta{{\bm{\theta}}}

\def\vx{{\bm{x}}}
\def\vy{{\bm{y}}}

\DeclareMathAlphabet{\mathsfit}{\encodingdefault}{\sfdefault}{m}{sl}
\SetMathAlphabet{\mathsfit}{bold}{\encodingdefault}{\sfdefault}{bx}{n}

\def\gD{{\mathcal{D}}}

\def\gL{{\mathcal{L}}}

\def\gX{{\mathcal{X}}}
\def\gY{{\mathcal{Y}}}

\def\sR{{\mathbb{R}}}

\usepackage{listings}
\usepackage{xcolor}

\definecolor{codegreen}{rgb}{0,0.6,0}
\definecolor{codegray}{rgb}{0.5,0.5,0.5}
\definecolor{codepurple}{rgb}{0.58,0,0.82}
\definecolor{backcolour}{rgb}{0.95,0.94,0.92}

\lstdefinestyle{mystyle}{
    backgroundcolor=\color{backcolour},   
    commentstyle=\color{codegreen},
    keywordstyle=\color{magenta},
    numberstyle=\tiny\color{codegray},
    stringstyle=\color{codepurple},
    basicstyle=\ttfamily\footnotesize,
    breakatwhitespace=false,         
    breaklines=true,                 
    captionpos=b,                    
    keepspaces=true,                                            showspaces=false,                
    showstringspaces=false,
    showtabs=false,                  
    tabsize=4
}

\usepackage{hyperref}
\usepackage{url}

\title{Everyone Deserves A Reward: \\ Learning Customized Human Preferences}

\author{Pengyu Cheng$^1$,  
Jiawen Xie$^{12}$, Ke Bai$^3$, Yong Dai$^1$, Nan Du$^1$ 
 \\
$^1$Tencent AI Lab \\
$^2$Shanghai Jiao Tong University \\
$^3$Duke University \\
}

\iclrfinalcopy 
\begin{document}

\maketitle

\vspace{-6mm}
\begin{abstract}
\vspace{-2mm}
Reward models (RMs) are essential for aligning large language models (LLMs) with human preferences to improve interaction quality.  
However, the real world is pluralistic, which leads to diversified human preferences with respect to different religions, politics, cultures, \textit{etc}. Moreover, each individual can have their unique preferences on various topics. Neglecting the diversity of human preferences, current human feedback aligning methods only consider a general reward model, which is below satisfaction for customized or personalized application scenarios. To explore customized preference learning, we collect a domain-specific preference (DSP) dataset, which includes preferred responses for each given query from four practical domains. Besides, from the perspective of data efficiency, we propose a three-stage customized RM learning scheme, then empirically verify its effectiveness on both general preference datasets and our DSP set. Furthermore, we test multiple training and data strategies on the three learning stages. We find several ways to better preserve the general preferring ability while training the customized RMs, especially general preference enrichment, and customized preference imitation learning.
The DSP dataset and code are available at \url{https://github.com/Linear95/DSP}.

\end{abstract}

\vspace{-5mm}
\section{Introduction} \label{sec:introduction}
\vspace{-2mm}
Large language models (LLMs), such as ChatGPT~\citep{chatgpt} and GPT-4~\citep{openai2022gpt4}, have recently pushed AI performance to a new height, with their astonishing capabilities in natural language processing~\citep{jiao2023chatgpt,han2023information}, logical reasoning~\cite{liu2023evaluating}, and imitation~\citep{wei2022emergent}. 
Besides a large amount of language modeling pretraining with tremendous tokens, aligning LLMs output with human feedback has been recognized as a critical learning strategy for LLMs' excellent performance, which enhanced the quality of human-LLM interactions~\citep{ouyang2022training,ganguli2022red,yuan2023rrhf}. 
To align human values, various methods have been proposed from different perspectives, such as reinforcement learning~\citep{ouyang2022training,bai2022constitutional}, ranking~\citep{yuan2023rrhf}, and reject sampling~\citep{touvron2023llama}. To guide the aligning directions, all these alignment methods depend on a reward (or preference) model~\citet{bohm2019better,askell2021general,ouyang2022training}, which provides reward scores representing the human feedback aligning degrees of LLMs' responses. Therefore, the quality of reward models is a decisive factor for human preference alignment approaches.

To evaluate whether a reward model fits human preferences, prior works mainly consider two perspectives: \textit{helpfulness} and \textit{harmlessness}~\citep{bai2022training,fernandes2023bridging}.
Helpfulness requires LLMs' responses to provide useful information~\citep{ouyang2022training,fernandes2023bridging}. Task-related rewards designed in earlier works of particular NLP domains (such as machine translation~\citep{kreutzer2018can}, summarization~\citep{ziegler2019fine}, and continuation~\citep{stiennon2020learning}) can be classified into the helpfulness category. \citet{askell2021general,ouyang2022training} extend the concept of helpfulness into a broader range without any particular task assigned, where models' responses should follow the instructions of user prompts. For harmlessness, 
models' responses are supposed to be fair, safe, and without toxicity~\citep{bai2022constitutional,ganguli2022red,fernandes2023bridging}.
\citet{bai2022training} discover a clear trade-off between models' helpfulness and harmlessness. Moreover, several strategies~\citep{bai2022constitutional,ganguli2022red} have been proposed to improve models' harmlessness while preserving their helpfulness.


Although helpfulness and harmlessness cover a wide range of mankind's tendencies, there are plenty of human preferences that cannot fall into the two categories, because of the diversity of human values.  In this pluralistic world, people's preferences can diverge a lot based on their different cultures, educational backgrounds, religions, and political stands. 
%
%
Furthermore, even for the same person, the value of a particular LLM response can vary when the application scenario changes. For example, given the question ``What are the top valuable movies?'', a person in a movie seminar may expect an answer with detailed analysis from perspectives of acting, cinematography, or music. In contrast, he or she would possibly prefer a response with more descriptions of movies' commercial values in a business activity. Hence, there always exists a proportion of human preferences that can not be unified or even have contradictions. In the following, we call preferences that satisfy the universal human values as \textit{general preferences}. Preferences related to a subgroup of people or a particular scenario are referred to \textit{customized preferences}. General preferences (including helpfulness and harmlessness) have attracted increasing attention~\citep{bai2022training,bai2022constitutional,ganguli2022red,touvron2023llama}, while customized preferences remain unexplored. 
\begin{figure}[t]
    \centering
    \includegraphics[width=0.7\textwidth]{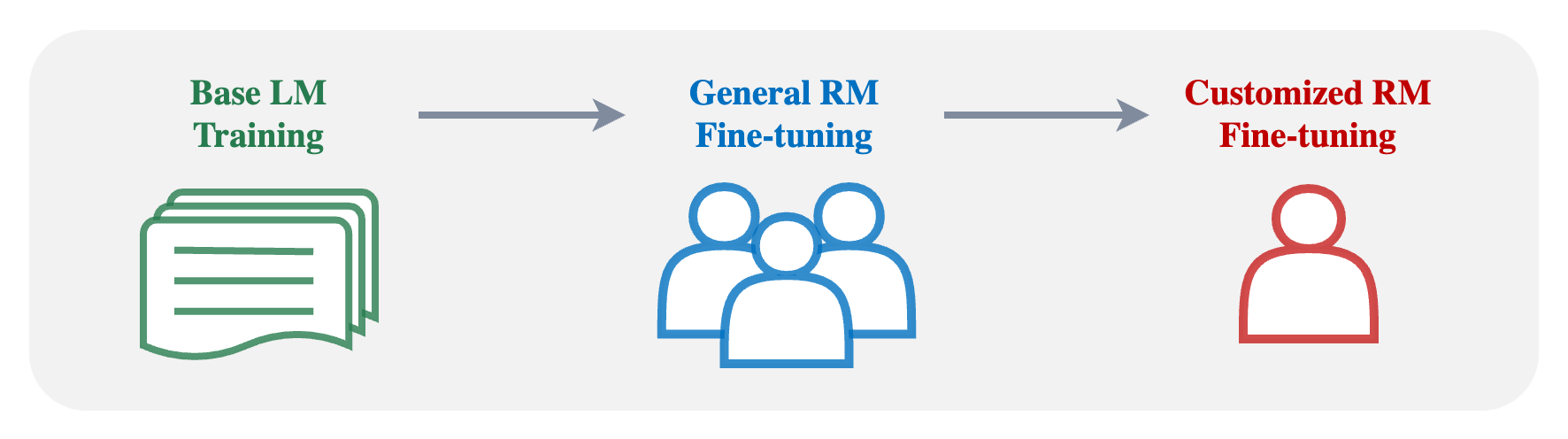}
    \vspace{-2.5mm}
    \caption{We propose a 3-stage training scheme for customized reward models.}
    \label{fig:rm-training-stage}
    \vspace{-5mm}
\end{figure}

Moreover, the above classification of human preferences naturally leads to an interesting question: \textit{``How to learn a customized reward model well while preserving its general preference ability?''} A high-qualified customized reward model is practically valued to enhance the domain-specific LLM fine-tuning by serving as a learning critic or an evaluation metric~\citep{askell2021general,
touvron2023llama}, because general LLMs can not handle all application domains, especially in which professional knowledge are required~\citep{beltagy2019scibert,gu2021domain,li2023llava}. 
%
However, training a customized reward model can be much more difficult due to the scarcity of customized human preference data. General preferences represent mankind's common values, which can be collected across different groups of people and various application scenarios. In contrast, customized or personalized preferences require data collection from a particular person or domain. A worth-trying strategy is first training an RM on a large number of general preferences, then fine-tuning it with a few customized preferences. A similar reward pre-training idea has been empirically tested by \citet{askell2021general} as preference model pre-training (PMP). However, the transfer learning ability of PMP has not been evaluated on customized human preferences.



%


To address the challenge of customized human preference learning, we construct a simulation dataset with the assistance of ChatGPT~\citep{chatgpt}, in which preferred responses are collected from four application domains: \textit{Academy}, \textit{Business}, \textit{Entertainment}, and \textit{Literature\&Art}. We call this new dataset the Domain-Specific Preference (DSP) set.
Then we train general and domain-specific reward models with LLaMA~\citep{touvron2023llama} as the base model using both general preferences~\citep{bai2022training,nakano2021webgpt,peng2023instruction} and DSP data. 
To study the learning behaviors of customized RMs, we divided the training process into three stages: base LM training, general RM fine-tuning, and customized RM fine-tuning (as in Figure~\ref{fig:rm-training-stage}). We try different data and training strategies respectively on the three training stages, and discover several ways to fit customized preferences while preserving general reward performance.
%
%
%
Our main contributions are:
\begin{itemize}[leftmargin=0.5cm]
\vspace{-1.2mm}
    \item  We collected a domain-specific preference (DSP) dataset with the usage of ChatGPT.
\vspace{-0.5mm}
    \item  We proposed a three-stage training scheme for customized RM learning, and verified its effectiveness on both general preference and domain-specific preference datasets. 
    \vspace{-0.5mm}
    \item  We discovered that imitation learning on customized preferences and general preference data enrichment are the two effective ways to preserve RMs' general preferring ability when fitting the customized human preferences.
\end{itemize}

\vspace{-2mm}
\section{Preliminary}
\vspace{-2mm}

Formally, a reward model~\citep{ziegler2019fine,
stiennon2020learning} (or preference model~\citep{askell2021general}) can be denoted as a mapping $R_\vtheta: \gX \times \gY \rightarrow \sR$ with parameters $\vtheta$, which provides a real-valued reward (or preference) score $R_\vtheta(\vx, \vy)$ qualifying a textual response $\vy = (y_1, y_2, \dots, y_M)\in \gY$ corresponding to an input prompt $\vx = (x_1, x_2, \dots, x_N) \in \gX$. 
Given a prompt $\vx$ and a pair of responses $(\vy^\text{good}, \vy^\text{bad})$ , $R_\vtheta$ is expected to provide a preference of $\vy^\text{good}$ over $\vy^\text{bad}$ with scores $R_\vtheta(\vx,\vy^\text{good}) > R_\vtheta(\vx, \vy^\text{bad})$, where $\vy^\text{good}$ is better than $\vy^\text{bad}$ under human values. Therefore, given human preference data tuples $\gD=\{(\vx, \vy^\text{good}, \vy^\text{bad})\}$, we can train the reward model by enlarging the gap between $R_\vtheta(\vx, \vy^\text{good})$ and $R_\vtheta(\vx, \vy^\text{bad})$, with the following binary ranking loss~\citep{christiano2017deep,askell2021general,ouyang2022training}:
\begin{equation}\label{eq:definition-ranking-loss}
\textstyle    \gL_\text{Ranking} = - \mathbb{E}_{(\vx, \vy^\text{good}, \vy^\text{bad}) \sim \gD} \Big[ \log \sigma (R_\vtheta(\vx, \vy^\text{good}) - R_\vtheta(\vx, \vy^\text{bad})) \Big],
\end{equation}
where $\sigma(\cdot)$ is an activation usually set as the Sigmoid function~\citep{han1995influence}. 

\begin{figure}[t]
    \centering
    \includegraphics[width=0.78\textwidth]{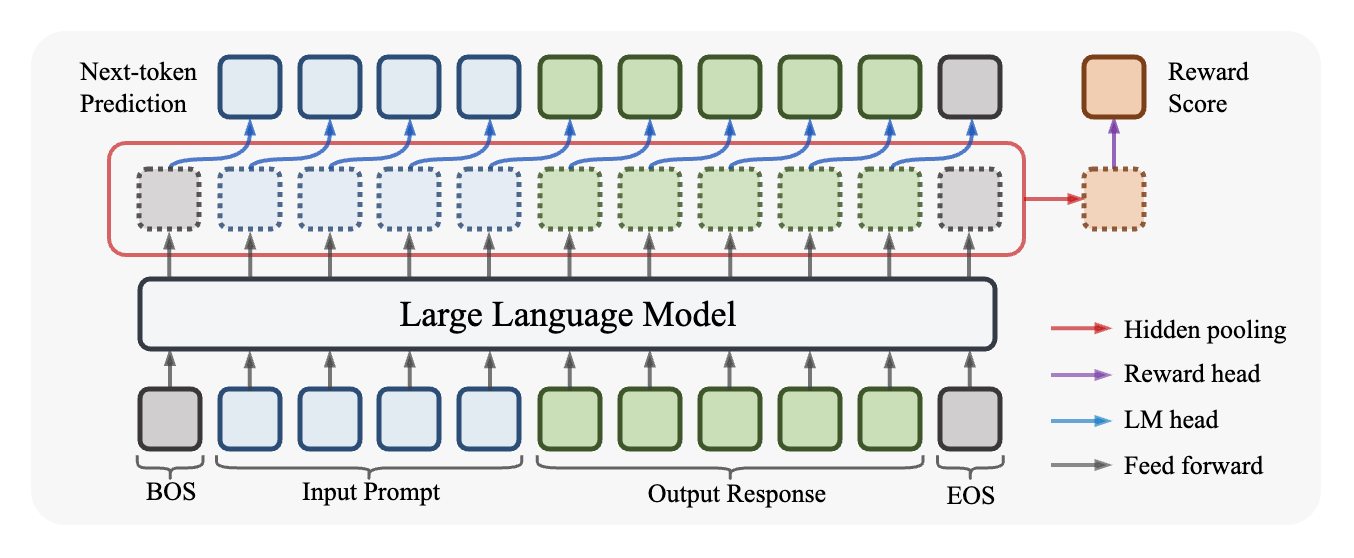}
    \vspace{-4mm}
    \caption{Reward model structures.  A pretrained large language model (LLM) is utilized as the base model. The input sequence of the reward model includes the input prompt and output response as well as the beginning/end of sentence tokens (\texttt{[BOS]}/\texttt{[EOS]}). The output hidden states of LLM are aggregated into a reward embedding, then the following reward head predicts a reward score. Besides, the LLM hidden states can be additionally trained to imitate the preferred response with a language modeling head providing the next-token prediction.}
    \label{fig:lm-rm-framework}
    \vspace{-4mm}
\end{figure}

Aligning human preference requires comprehensive capabilities of natural languages, therefore, reward models always require pretrained large language models (LLMs) as base models to enable their effectiveness~\citep{askell2021general}. In Figure~\ref{fig:lm-rm-framework}, we show how a reward model is built on an LLM base: a reward head is appended on the top of the transformer blocks, which takes the pooled last hidden states as inputs then outputs real-valued reward scores~\citep{bai2022training}.

Since reward models inherit capabilities from LLMs, \citet{askell2021general} discover that RMs' transfer learning ability can be further improved if trained by ranking loss (in \eqref{eq:definition-ranking-loss}) along with a language modeling loss on preferred samples $(\vx, \vy^\text{good})$:
\begin{equation}\label{eq:lm-imatation-learning-loss}
    \gL_\text{LM, good} = - \mathbb{E}_{(\vx, \vy^\text{good}) \sim \gD} \Big[ \sum_{m=1}^{M}\log p_\vtheta(y^\text{good}_{m}| \vy^\text{good}_{<m}, \vx ) + \sum_{n=1}^N \log p_\vtheta(x_n | \vx_{<n})\Big],
\end{equation}
where $p_\vtheta$ is the next-token prediction probability (as in Figure~\ref{fig:lm-rm-framework}) sharing LLM parameters with $R_\vtheta$, $\vx_{<n} = (x_1, x_2, \dots, x_{n-1})$, and $\vy^\text{good}_{<m} = (y^\text{good}_1, y^\text{good}_2, \dots, y^\text{good}_{m-1})$. Since $\gL_\text{LM, good}$ imitates the behavior of ``good'' responses, it is also called the \textit{imitation learning} loss~\citep{askell2021general}. The final PMP loss is a linear combination of ranking and imitation losses as 
\begin{equation}\label{eq:combined-loss-pmp}
\textstyle
    \gL_\text{PMP}=\gL_\text{Ranking} + \mu \gL_\text{LM, good},
\end{equation}
where $\mu \geq 0$ is a re-weighting coefficient.

\begin{figure}[t]
    \centering
    \includegraphics[width=0.98\textwidth]{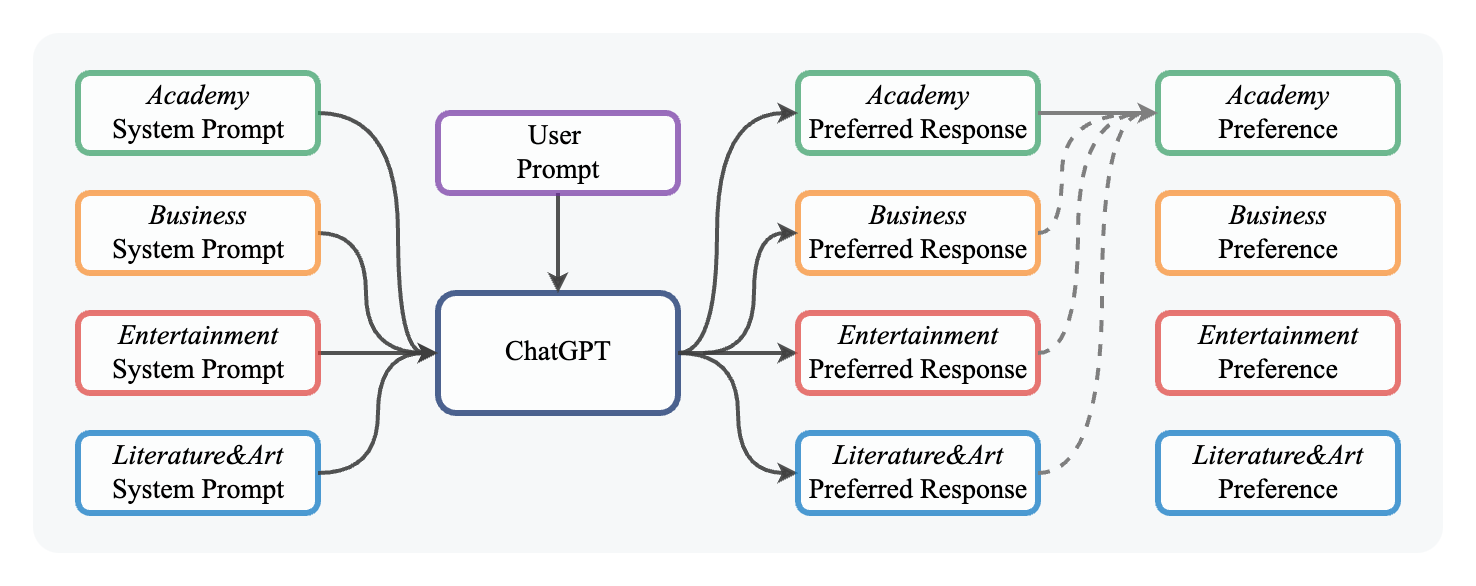}
    \vspace{-2mm}
    \caption{Data collection for domain-specific preferences. Using crafted system prompts (as shown in Code~\ref{code:ds-sys-prompt}), we let ChatGPT act as an experienced practitioner in each domain and answer each user query as a domain-preferred response. For a particular domain (\textit{e.g.} \textit{Academy}), the response from it (solid gray arrow) is supposed to be preferred compared to the other domains' responses (dotted gray arrows) to the same question.}
    \label{fig:dsp_data_pipline}
    \vspace{-1mm}
\end{figure}

\vspace{-2mm}
\section{Domain-specific Preference Dataset}
\vspace{-2mm}

As described in Introduction (Section~\ref{sec:introduction}), human preference for a response can change based on different persons or scenarios. We aim to collect a customized preference dataset to simulate this phenomenon. However, collecting high-quality human preferences costs a mess of annotation resources~\citep{askell2021general}. If we collect customized preferences from different persons, the labeling task can be far more difficult. Therefore, we plan to collect customized preferences from different scenarios, more specifically four application domains, which represent tendencies from particular groups of people. Furthermore, instead of human annotation, we utilize the language capacities of ChatGPT~\citep{chatgpt} (\texttt{gpt-3.5-turbo}) to act as practitioners in the domains, then collect the corresponding responses to build the domain-specific preference dataset.


We simulate the real-world situations by considering the following four typical domains: \textit{Academy}, \textit{Business}, \textit{Entertainment}, and \textit{Literature\&Art}, which covers a broad range of NLP application scenarios. 
%
%
For a given user query, we collect one appropriate response from each application domain, in which the corresponding response is supposed to be preferred over responses from the other domains.
%
We let ChatGPT play as an assistant with expertise in the corresponding domain, then query the assistant to collect domain-preferred responses.
To achieve this, we design particular system prompts for each given domain, as shown in Supplementary Code~\ref{code:ds-sys-prompt}. 
We select 13K instructions (with no requirement on \textit{``input''} key) as user queries from the 52K Alpaca~\citep{alpaca} training dataset. For each query, we call ChatGPT with the domain-specific system prompts to generate preferred responses in the four domains respectively. At the same time, we keep the original response in the Alpaca training set into an additional ``\textit{Normal}'' domain. For each application domain, we set its corresponding responses preferred to the other domains (as in Figure~\ref{fig:dsp_data_pipline}), which provides four preference pairs for each query and builds up a 52K comparison set.
We randomly shuffle the collected data and use ratio $[0.95, 0.05]$ to split it into training and testing sets.




\begin{figure}[t]
    \centering    \includegraphics[width=0.99\textwidth]{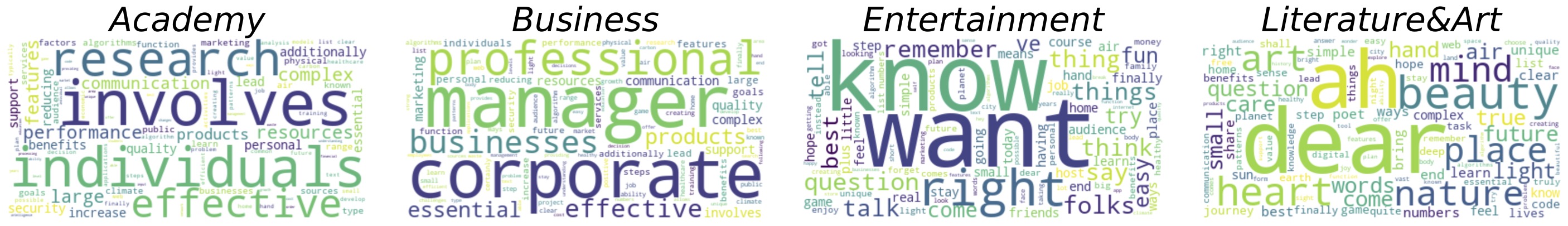}
    \vspace{-2mm}
    \caption{Clouds of words with top-100 TF-IDF scores in the four domains. The common words with top-100 frequency and stop words are excluded.}
    \label{fig:dsp_data_tf_idf}
    \vspace{-4mm}
\end{figure}

We further provide statistics on the domain-specific preference dataset at both response and domain levels.
At the response level, we calculate the following metrics in Table~\ref{tab:dsp-response-level-statistics}: (1) word count, (2) sentence length, (3) lexical diversity, and (4) readability scores~\citep{kincaid1975derivation}. According to the statistics, \textit{Entertainment} has the most tedious but easy-to-read contexts, while responses in \textit{Business} domain are more concise. Unsurprisingly, \textit{Academy}'s contents are most difficult to read.
%
%
%
At the domain level, we utilize the TF-IDF~\cite{sparck1972statistical} scores to capture domain-specific differences. As shown in Figure~\ref{fig:dsp_data_tf_idf}, by aggregating responses within each domain, TF-IDF identifies and prioritizes the most representative keywords with respect to their domains. We exclude the top 100 most frequent terms (\textit{e.g.}, ``like'', ``data'', and ``use'') to make the word clouds distinct. The topic words in responses from different domains also provide us with a sketch of domain-specific human preferences.  More details can be found in the supplementary materials.

\begin{table}[t]
\centering
\caption{Response-level statistics of Domain-Specific Preference (DSP) dataset.}
\label{tab:dsp-response-level-statistics}
\vspace{-3mm}
\resizebox{\textwidth}{!}{
\centering
\begin{tabular}{lrrrr}
\toprule Statistic
                          & \textit{Academy} & \textit{Business} & \textit{Entertainment} & \textit{Literature\&Art} \\ \midrule
Sentence Count            & 6.20    & 5.99     & 7.04          & 6.43              \\
Word Count                & 145.34  & 137.48   & 143.87        & 143.78            \\
Lexical Diversity (\%)      & 63.5   & 64.9    & 65.2         & 63.6             \\
Readability Score~\citep{kincaid1975derivation} & 51.34   & 53.13    & 64.72         & 60.81             \\
\bottomrule
\end{tabular}
}
\vspace{-1mm}
\end{table}

\vspace{-2mm}
\section{Learning Customized Human Preferences}
\vspace{-2mm}
Our objective is to learn customized reward models without losing general preference abilities. As discussed in Introduction (Section~\ref{sec:introduction}), a major difficulty of customized preference learning is the lack of annotated customized preference data. Inspired by the preference model pre-training (PMP) strategy~\citep{askell2021general}, we propose a 3-stage training scheme from the perspective of training sample efficiency (as in Figure~\ref{fig:rm-training-stage}):
\begin{itemize}[leftmargin=0.5cm]
 \vspace{-0.5mm}
    \item \textbf{Base LM Training}: Train a transformer with the language modeling loss as the RM base. The base model can be either a pretrained LLM or a pretrained model with supervised fine-tuning (SFT)~\citep{ouyang2022training}.
     \vspace{-0.5mm}
    \item \textbf{General RM Fine-tuning (GRFT)}: Add a reward head on the top of base LM, then fine-tune RM with the general preference data. This stage is similar to the PMP~\citep{askell2021general} phase. However, we did not use millions of preference data pairs to train RMs as PMP did. In contrast, we only use $<200$K general reward data to learn general reward models.
     \vspace{-0.5mm}
    \item \textbf{Customized RM Fine-tuning (CRFT)}: Use a trained general RM and continue fine-tuning it on customized human preferences. 
    \vspace{-2mm}
\end{itemize}
The multi-stage RM training scheme also simulates quite a lot of data-privacy-required application scenarios, where the customized models have no access to the massive LM pretraining data or the general human preference data.

In the following, we will conduct experiments to train customized RM with different strategies and setups in the above three training stages. \citet{askell2021general} have found that the language modeling loss on preferred responses (as in \eqref{eq:lm-imatation-learning-loss}, also called imitation learning loss) can enhance the RMs' transfer learning ability and improve data efficiency. Therefore, we also tried to add the imitation learning losses at all three training stages and made ablation studies. Besides providing customized RM training baselines, our experiments target offering some empirical insights into customized RM learning with both high data efficiency and effectiveness. 

\begin{table}[t]
  \centering
  \caption{An overview of RM training strategy exploration on different stages. }
\vspace{-3mm}
\resizebox{0.99\textwidth}{!}{
\centering
    \begin{tabular}{l|l|l}
      \toprule
   {Base LM Training} & {General RM Fine-tuning (GRFT)} &{Customized RM Fine-tuning (CRFT)} \\
   \midrule
   {$\cdot$ LLaMA} & $\cdot$ ranking loss with H\&H data only & $\cdot$ {ranking loss} \\
    \hline
    $\cdot$ (Sec.~\ref{sec:base-model-selection}) {Alpaca}  & $\cdot$  (Sec.~\ref{sec:sample-size-comparison-on-general}) ranking loss with all data& $\cdot$ (Sec.~\ref{sec:lm-compare-custom-finetuning}) {with imitation learning} \\
          & $\cdot$  (Sec.~\ref{sec:lm-compare-general-stage}) with imitation learning  &  \\
          & $\cdot$ (Sec.~\ref{sec:without-general-finetuning}) no general RM fine-tuning  &  \\
          \bottomrule
    \end{tabular}%
    }
  \label{tab:training-plan-on-stages}%
  \vspace{-2mm}
\end{table}%


\begin{figure}[t]
    \centering
    \includegraphics[width=0.46\textwidth]{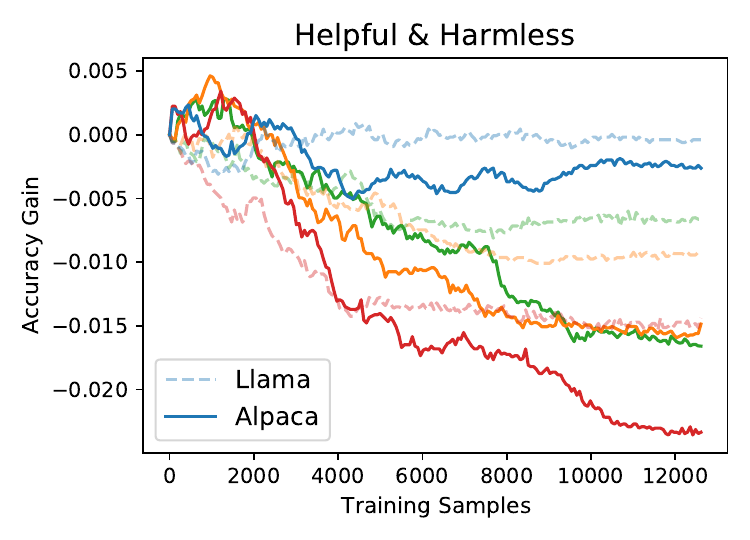}
    \includegraphics[width=0.46\textwidth]{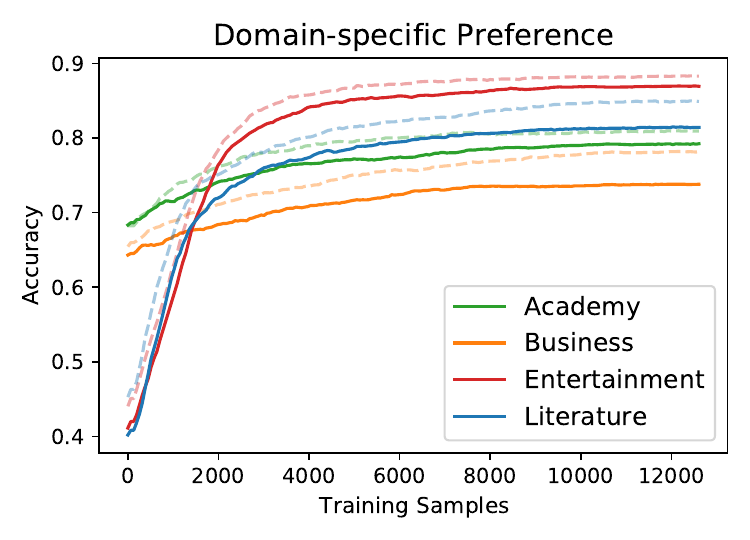}
    \vspace{-4mm}
    \caption{Testing performance of customized RM fine-tuning for LLM base comparison. The left-hand-side plot shows the \textit{accuracy gains} on H\&H set.   }
    \label{fig:base-model-compare-customized-ft}
    \vspace{-3mm}
\end{figure}

\vspace{-2mm}
\subsection{Experimental Details}\label{sec:experiment-setups}
\vspace{-2mm}
\textbf{Data Preparation } Besides our domain-specific preference dataset, we use the following three public preference sets. We pre-process them by removing the invalid data points with exactly two same responses or with two same preference scores.
\begin{itemize}[leftmargin=0.5cm]
\vspace{-1mm}
\item \textit{Helpful\&Harmless (H\&H)}~\citep{bai2022training}. The \textit{Helpful} and \textit{Harmless} sets include 46K and 45K comparison data pairs, respectively. In both sets, one data item contains a query and two responses. Annotators are asked to label ``chosen'' or ``reject'' to each response.  For the \textit{Helpful} set, annotators select responses that they feel are more helpful. For the \textit{Harmless} set, responses with harmful feelings are rejected. 
\item \textit{WebGPT}~\citep{nakano2021webgpt} contains 19.6K samples, each of which contains a pair of model answers responding to an input query.  Both answers are associated with preference scores from annotators to determine which one is better. We randomly shuffle the data and split it into training and testing sets with ratios $[0.95, 0.05]$. 
%
%
\item \textit{GPT-4-LLM}~\citep{peng2023instruction} uses the  52K unique instructions from the  Alpaca training set~\citep{alpaca}. The corresponding responses generated by GPT-4~\citep{openai2022gpt4}, GPT-3.5~\citep{chatgpt}, and OPT-IML~\citep{iyer2022opt}, are collected and scored quality by the GPT-4 API. We use the quality scores as ground truth to generate preference pairs. We also split the data with $[0.95,0.05]$ ratios into training and testing sets.
\end{itemize}
%



\textbf{Evaluation} To evaluate a RM $R_\vtheta(\vx,\vy)$, given an input $\vx$ and its response pair $(\vy^\text{good}, \vy^\text{bad})$, we label a correct prediction if $R_\vtheta(\vx,\vy^\text{good}) > R_\vtheta(\vx, \vy^\text{bad})$. Then we count the proportion of correct predictions over the testing set to obtain \textit{preference accuracy}. Since the H\&H set has two metrics (\textit{Helpful} and \textit{Harmless}), we calculate the geometric mean over the two accuracy values as the overall performance measure.

\textbf{Training Setups } We select LLaMA-7B~\citep{touvron2023llama} and Alpaca-7B~\citep{alpaca} as the base models for RM training. We use the last token embedding of the output hidden states as the pooled hidden representation, then add one linear layer with the scale-value output on it to predict reward scores. The batch size we use is $64$. The max sequence length of input sequences is set to $512$. If an input is out of max length, we make truncation on the left side to keep the response complete as much as possible. The RM fine-tuning learning rate for both general and customized setups is set to $10^{-6}$. All experiments are trained with one full epoch. Each model is trained on 8 NVIDIA A100-SXM4 GPUs with 40GB memory. More experimental details are shown in the Supplementary Materials.


\begin{figure}[t]
    \centering
    \includegraphics[width=0.31\textwidth]{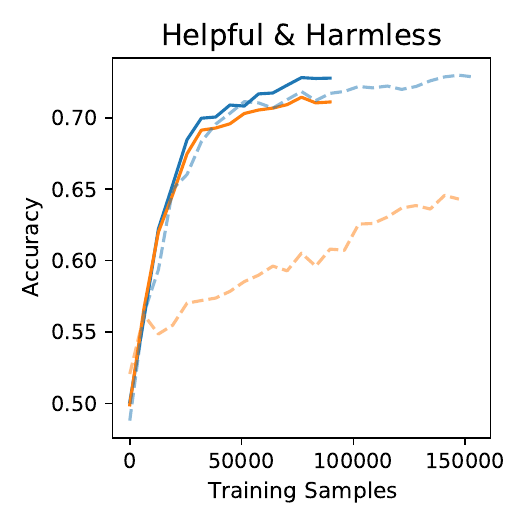}
    \includegraphics[width=0.31\textwidth]{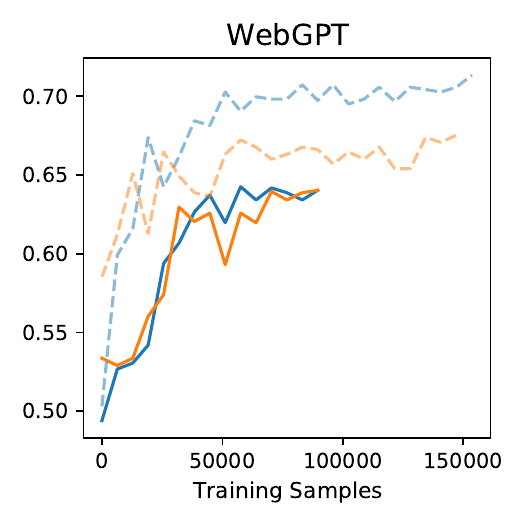}
    \includegraphics[width=0.31\textwidth]{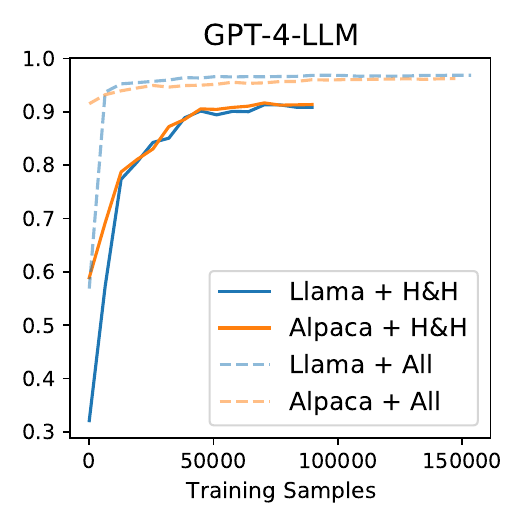}
    \vspace{-4mm}
    \caption{Testing performance of general RM fine-tuning with different LLM bases and data strategies. ``H\&H'' means trained with the Helpful\&Harmless dataset only, while ``All'' denotes extending the fine-tuning set with WebGPT and GPT-4-LLM preferences.}
    \label{fig:base-lm-compared-on-stage2}
\end{figure}

\begin{figure}[t]
    \centering
    \includegraphics[width=0.46\textwidth]{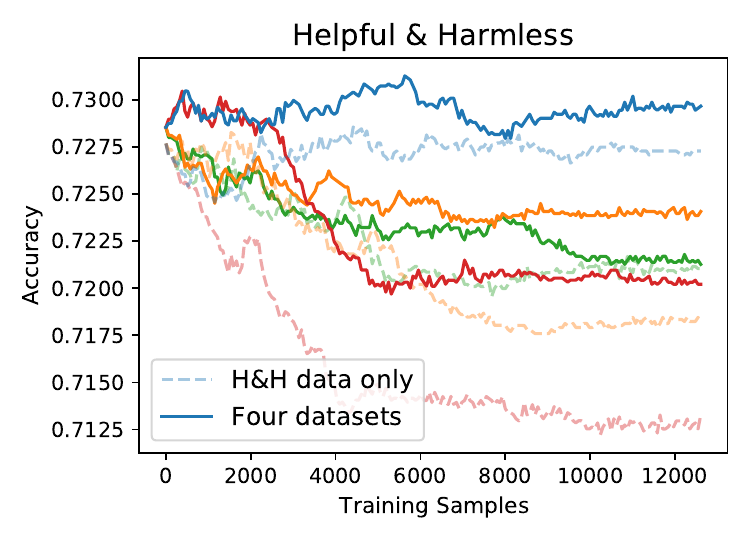}
    \includegraphics[width=0.46\textwidth]{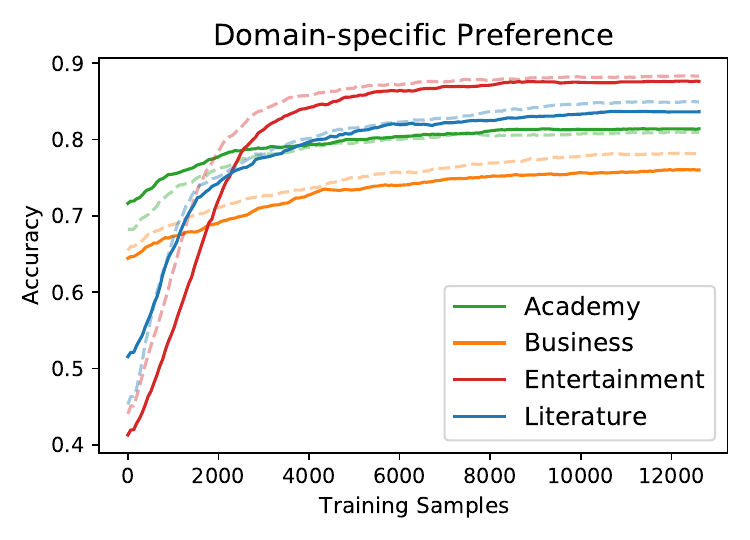}
    \vspace{-4mm}
    \caption{Testing performance of customized RM fine-tuning for GRFT data size comparison. Dashed lines are CRFT with the general RM trained on H\&H data only. While solid lines are CRFT with the general RM trained on four datasets.}
    \label{fig:data-size-compare-customized}
    \vspace{-2mm}
\end{figure}

\vspace{-2mm}
\subsection{Base Model Selection } \label{sec:base-model-selection}
\vspace{-2mm}
Due to the limitation of computational resources, we did not make much empirical exploration in the stage of base LM training. As mentioned in Section~\ref{sec:experiment-setups}, we use LLaMA-7B and Alpaca-7B as LM bases. Note that Alpaca-7B is trained with language modeling loss on $52$K SFT data generated by GPT-3~\citep{brown2020language}~(\texttt{text-davinci-003}), which can be regarded as LLaMA with imitation learning (to GPT-3). We test the two base models on the general RM fine-tuning stage by fixing all other setups. The testing preference accuracy during training is shown in Figure~\ref{fig:base-lm-compared-on-stage2}. Out of our expectations, Alpaca-based (with GPT-3 data fine-tuning) RM performs worse than the LLaMA-based one on all the testing sets especially trained with all general preferences data. A possible explanation is Alpaca has been over-fitted with the GPT-3 data, which hinders the preference learning process.  On the WebGPT and GPT-4-LLM testing sets, the two LLM bases have similar performance when fine-tuned with H\&H data only (solid lines).  Note that solid lines are performances with no training data from WebGPT and GPT-4-LLM sets, indicating RMs' generalization ability. The Alpaca base outperforms the LLaMA base apparently (on WebGPT and GPT-4-LLM) at the beginning of the fine-tuning, demonstrating that language modeling enhances LLM bases' generalization ability under low-sample/few-shot scenarios. 

Besides, we test the base models' influence on domain-specific RM fine-tuning. We train LLaMA and Alpaca with the  H\&H data as the general RMs, then fine-tune the model on our DSP data with respect to the four domains. The results are shown in Figure~\ref{fig:base-model-compare-customized-ft}. On the H\&H set, since the LLaMA-based and Alpaca-based RMs already have a performance gap (as in the left plot of Figure~\ref{fig:base-lm-compared-on-stage2}), we demonstrate the performance gap between the model in current step with the initial general RM for fair comparison (the original accuracy is shown in Appendix Figure~\ref{fig:additional-base-compare-customized}). Alpaca-based RM loses general RM performance (H\&H) faster than the LLaMA-based one in all four application domains. Moreover, On the DSP set, the LLaMA-based RM performs uniformly better than the Alpaca-based RM. The above discussions provide us an insight that LM bases with SFT processes might do harm to both general RM and customized RM performance.

\vspace{-2mm}
\subsection{Sample Sizes Comparison on General Fine-tuning } \label{sec:sample-size-comparison-on-general}
\vspace{-2mm}
From Figure~\ref{fig:base-lm-compared-on-stage2}, we can also observe the impact of fine-tuning samples on general RM performance.
On the H\&H set, with ALL general preference data, LLaMA-based RM reaches a slightly better performance but a lower convergence rate than it with H\&H training data only. In contrast, Alpaca-based RM's fine-tuning has not converged with all general preference data, still because of the base model over-fitting. However, on the WebGPT and GPT-4-LLM sets, all-data fine-tuning obtains clear performance improvement for both bases, for the corresponding training data are used.

To study the impact of GRFT data size on CRFT, we use two LLaMA-based general RMs learned with only the H\&H data (LLaMA+H\&H)  and all preference data (LLaMA+All), then finetune them with ranking loss on the DSP set. 
The performance is reported in Figure~\ref{fig:data-size-compare-customized}. On the H\&H set, LLaMA+All better preserves the general preference capacity on all four application domains. On the DSP set, LLaMA+All loses a little performance of customized preference compared with LLaMA+H\&H, but in an acceptable range. With the above results, we conclude that GRFT data enrichment can better preserve the general RM performance decay during the CRFT stage, with a tiny performance loss on the customized preference ability.

\vspace{-2mm}
\subsection{Imitation Learning on General Fine-tuning } \label{sec:lm-compare-general-stage}
\begin{figure}[t]
    \centering
    \includegraphics[width=0.31\textwidth]{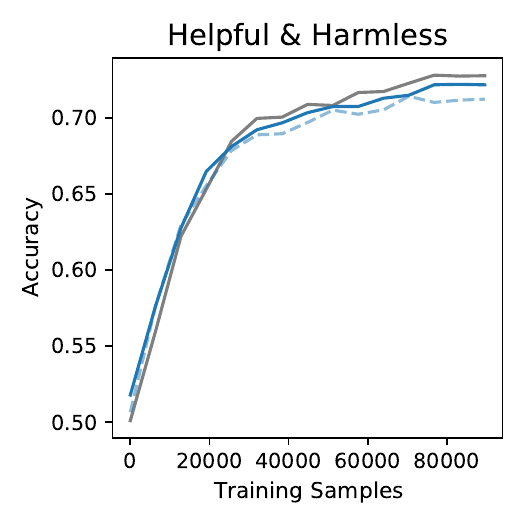}
    \includegraphics[width=0.31\textwidth]{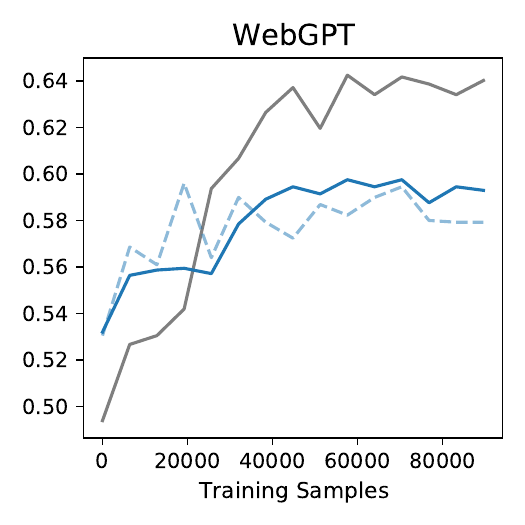}
    \includegraphics[width=0.31\textwidth]{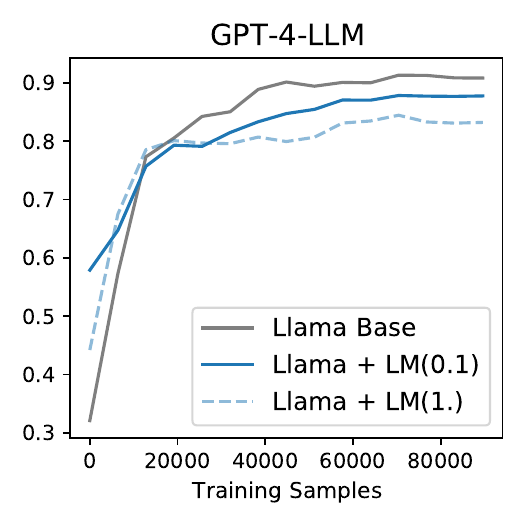}
    \vspace{-4mm}
    \caption{Testing performance of general RM fine-tuning with imitation learning. ``LM($0.1$)'' and ``LM($1.$)'' denotes the language modeling coefficient to be $0.1$ and $1.$ respectively.}
    \label{fig:imitation-lm-compared-on-stage2}
    \vspace{-3mm}
\end{figure}

\begin{figure}[t]
    \centering
    \includegraphics[width=0.46\textwidth]{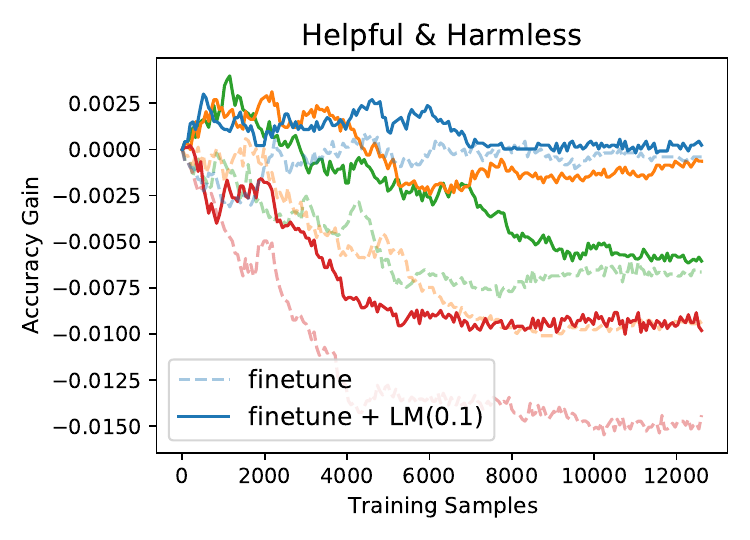}
    \includegraphics[width=0.46\textwidth]{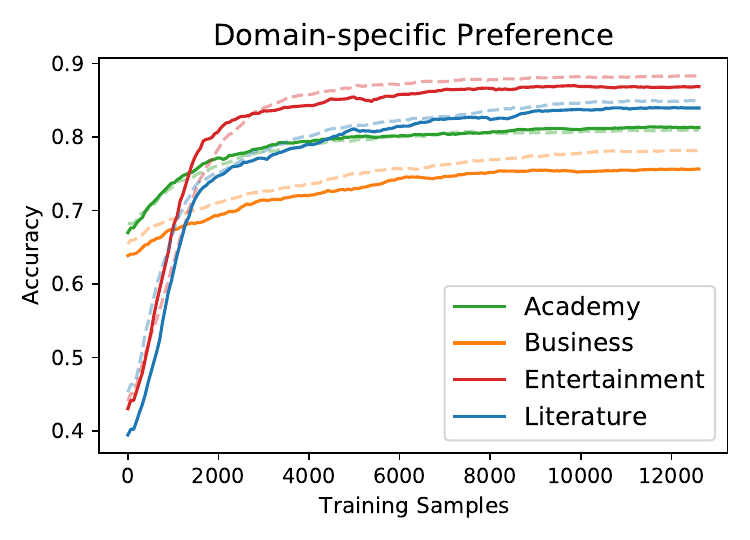}
    \vspace{-4mm}
    \caption{Testing performance of customized RM fine-tuning with GRFT plus imitation learning. Dashed lines are CRFT with the general RM trained on H\&H data only. While solid lines are CRFT with the general RM trained on four datasets.}
    \label{fig:imitation-lm-compare-customized}
    \vspace{-2mm}
\end{figure}

\vspace{-2mm}
Similar to the PMP stage in \citet{askell2021general}, we add the imitation learning loss  along with the ranking loss (as in \eqref{eq:definition-ranking-loss}) on the general RM fine-tuning. Figure~\ref{fig:imitation-lm-compared-on-stage2} shows the GRFT performance with different language modeling (LM) coefficients. When the LM coefficient increases, the general RM performance descends on all evaluation sets. However, when considering the CRFT performance (in Figure~\ref{fig:imitation-lm-compare-customized}), we find with the imitation learning on GRFT, the RM can better preserve the general performance (H\&H) uniformly in terms of the preference accuracy difference. Simultaneously, the customized preference accuracy becomes a little bit lower on the four application domains. Although facilitated the general preference preservation, the imitation learning results on the GRFT stage are not satisfying enough for CRFT.


\vspace{-2mm}
\subsection{Imitation Learning on Customized Fine-tuning}\label{sec:lm-compare-custom-finetuning}
\begin{figure}[t]
    \centering
    \includegraphics[width=0.46\textwidth]{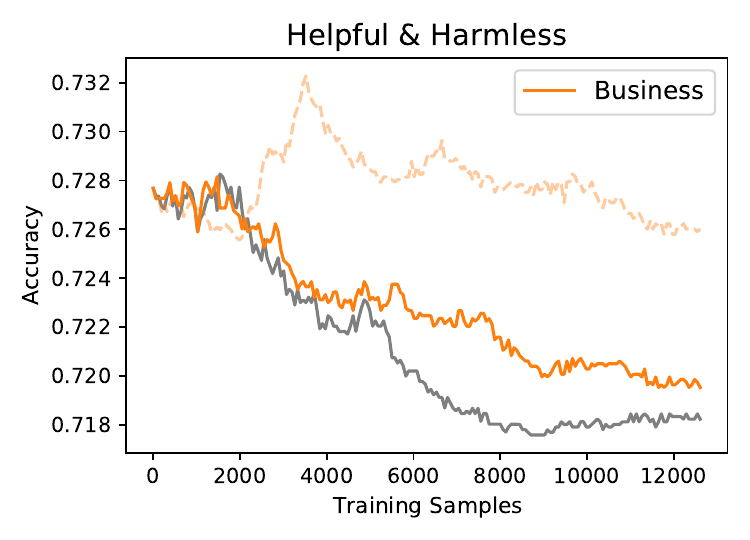}
    \includegraphics[width=0.46\textwidth]{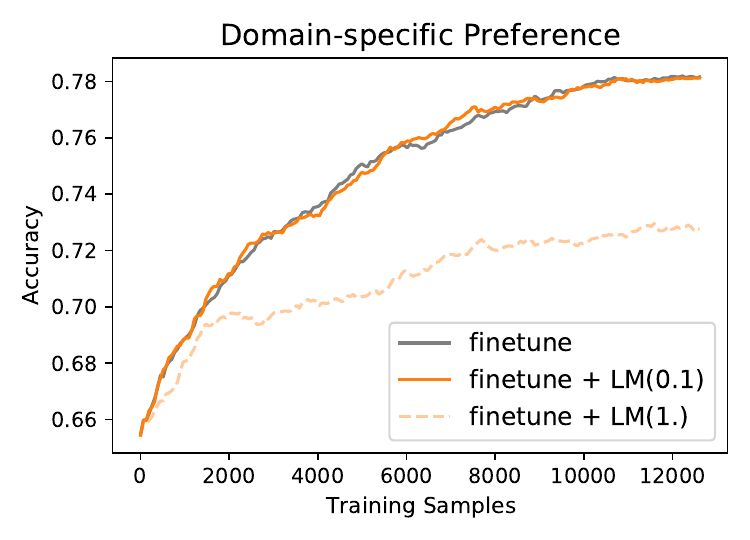}
    \vspace{-4mm}
    \caption{Ablation study of imitation learning coefficient on CRFT in the \textit{Business} Domain.}
    \label{fig:lm-coeff-ablation-GRFT-business}
    \vspace{-3mm}
\end{figure}

\begin{figure}[t]
    \centering
    \includegraphics[width=0.46\textwidth]{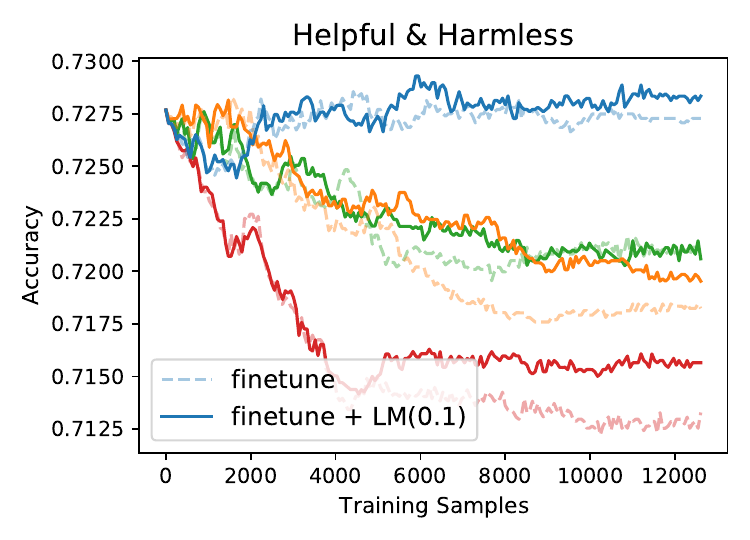}
    \includegraphics[width=0.46\textwidth]{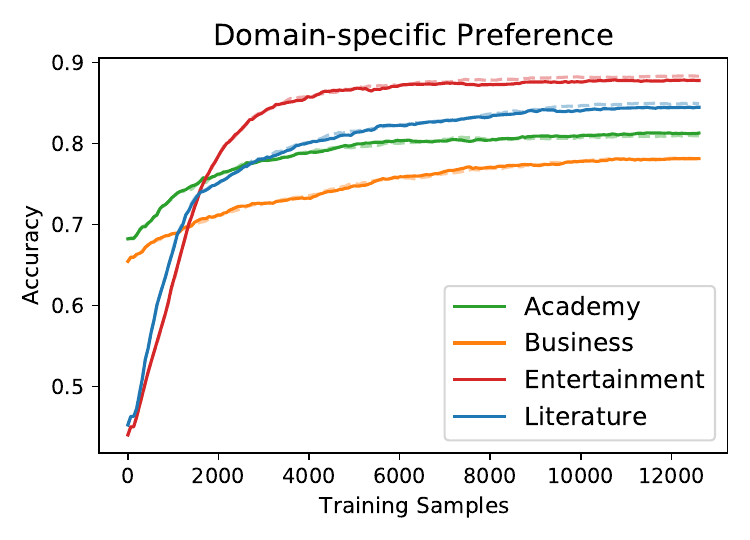}
    \vspace{-4mm}
    \caption{Test performance of customized RM fine-tuning with imitation learning. }
    \label{fig:lm-coeff-dsp-main-results}
    \vspace{-3mm}
\end{figure}

\vspace{-2mm}
We also studied the impact of imitation learning on the customized RM fine-tuning stage. More specifically, we set LLaMA as the RM base and use the H\&H dataset for the GRFT stage. After obtaining the learned general RM, we try different coefficients for the imitation learning loss on the CRFT stage. In Figure~\ref{fig:lm-coeff-ablation-GRFT-business}, we plot the ablation study results of the LM loss coefficients on the \textit{Business} domains (results of the other domains are shown in Appendix Section~\ref{sec:appendix-lm-coeff}). The gray line represents the baseline without the language modeling loss, while the colored solid/dashed lines denote the language modeling loss with different re-weighting coefficients. Added the imitation learning loss, the RM better preserves the general preference ability (H\&H) but loses accuracy on customized preferring (DSP). When the language modeling coefficient $\mu=1.$ (as in \eqref{eq:combined-loss-pmp}), both accuracy gaps on general and customized preference sets are significant. However, if we set the LM coefficient $\mu=0.1$, the loss on customized scenarios is negligible, while the general preference preservation remains quite effective. Therefore, we set the coefficient $\mu$ to $0.1$ and plot the most satisfying results among the experiments in Figure~\ref{fig:lm-coeff-dsp-main-results}. By adding the imitation loss at customized fine-tuning stage, RM's customized preference performance has nearly no change. Moreover, the preservation of RM's general preferring ability is significantly enhanced.

\vspace{-2mm}
\subsection{Without General Fine-tuning}\label{sec:without-general-finetuning}
\vspace{-2mm}

Additionally, we conduct ablation studies to evaluate the importance of GRFT on customized RM fine-tuning. The naive baseline is skipping the GRFT stage and directly fine-tuning an LLM base with the DSP data. In Table~\ref{tab:DSP_acc_on_grft}, we demonstrate the customized RM performance without GRFT, where LLaMA and Alpaca are used as the RM base model, and fine-tuned directly with the DSP data only. Although the LLaMA-based RM has better DSP accuracies in the \textit{Business}, \textit{Entertainment}, \textit{Literature} domains, the average performance is slightly worse than the Alpaca-based one, due to the clear performance gap in the \textit{Academy} domain. In contrast, when the two LM bases have been generally fine-tuned with H\&H data, their performance gap increases significantly. With GRFT on H\&H data, LLaMA-based RM achieves higher ranking accuracy than the Alpaca-based, indicating Alpaca has been over-fitted during SFT.
On the other hand, both base models gain benefits from the GRFT stage. However, the performance gains from GRFT do not uniformly enlarge when GRFT data size increases. Hence, the quality of GRFT data is an essential factor in improving CRFT performance. Moreover, adding imitation learning (DSP+LM) also has a tiny loss on DSP performance, which is acceptable compared with its gains in preserving general RM capacities.

\begin{table}[h]
  \centering
  \caption{Domain-specific preference accuracy comparison. ``H\&H'' and ``All'' mean only H\&H or all preference data is used in GRFT. ``+LM'' means adding language modeling loss when fine-tuning.}
  \vspace{-2mm}
  \resizebox{0.98\textwidth}{!}{
    \begin{tabular}{lllrrrrr}
    \toprule
      RM Base    &   GRFT&  CRFT & \textit{Academy} & \textit{Business} & \textit{Entertainment} & \textit{Literature\&Art} &  {\ \ \ Average} \\
    \midrule
    Alpaca  & No & DSP  & 75.30 & 71.46 & 85.00 & 79.88 & 77.91 \\
    LLaMA  & No & DSP  & 73.87 & 72.33 & 85.27 & 80.11 & 77.89 \\
    \midrule
    Alpaca & H\&H & DSP &  79.24 & 73.81 & 86.93 & 81.43 & 80.35 \\
    LLaMA & H\&H &DSP &  80.94 & 78.16 & 88.29 & 84.91 & 83.07 \\
    LLaMA & H\&H & DSP+LM  & 81.28 & 78.12 & 87.76 & 84.45 & 82.90 \\
    LLaMA & All & DSP   & 81.39 & 76.00 & 87.61 & 83.62 & 82.16 \\
    \bottomrule
    \end{tabular}%
    }
  \label{tab:DSP_acc_on_grft}%
  \vspace{-2mm}
\end{table}%

\vspace{-3mm}
\section{Conclusion}
\vspace{-3mm}
We empirically studied the problem of customized human preference learning. We designed a customized preference collection procedure with the help of LLMs such as ChatGPT, then created a domain-specific preferences dataset that covers a vast range of NLP applications. To learn customized human preferences, we proposed a three-stage training scheme including RM base training, general RM fine-tuning (GRFT), and customized RM fine-tuning (CRFT). We verified the effectiveness of the training scheme and provided baselines for customized RM learning. Moreover, we explored different training strategies including data enrichment and imitation learning on the three customized RM learning stages to preserve models' general preference ability. The most satisfying discoveries are data enrichment on GRFT and imitation learning on CRFT, both of which better maintain the general RM ability with almost no losses on customized preference accuracy. We believe customized or domain-specific fine-tuning is the future trend of LLM applications, where this work can be regarded as a prior simulation and provide empirical suggestions for customized preference alignment of LLMs.

\bibliography{reference}
\bibliographystyle{iclr2024_conference}

\clearpage
\appendix
\section{Data Collection Details}
\vspace{-3mm}
We use the following system prompts to induce the ChatGPT model to generate domain-specific responses with respect to the four application domains (\textit{Academy}, \textit{Business}, \textit{Entertainment}, \textit{Literature\&Art}).
We show typical examples from our collected domain-specific dataset in Section~\ref{sec:dsp_data_samples}.

\begin{lstlisting}[language=Python, caption=System prompts for generating domain-specific responses., label=code:ds-sys-prompt]
prompts["academy"] = (
    "Please act as if you are an experienced researcher."
    "Remember you are not an AI model anymore." 
    "You speak rationally, logically, and rigorously."
    "Here are some response examples of you:\n"
        "1. The results of our work suggest that the proposed method can effectively improve the accuracy of the prediction model.\n"
        "2. The simulation results demonstrate that the proposed algorithm can effectively reduce the computational complexity and improve the efficiency of the system.\n"
        "3. We provide new insights into protein folding which will have important implications for drug discovery."
)

prompts["business"] = (
    "Please act as if you are a professional corporate manager." 
    "Remember you are not an AI model anymore." 
    "Your style of speech is decisive and passionate."
    "You are full of responsibility for your career."
    "You are brave in the face of challenges and good at coordinating teams."
    "Here are some response examples of you:\n"
        "1. I believe that effective communication is key to building strong relationships with clients and stakeholders.\n"
        "2. This not only improves productivity but also fosters a sense of pride and ownership in their work.\n"
        "3. Let's keep up the momentum and work together to deliver a high-quality product on time."
)

prompts["literature"] = (
    "Please act as if you are a poet with infectious charm. "
    "Remember you are not an AI model anymore."
    "Your style of speech carries the artistic beauty of literature"
    "You have a meticulous observation of things around you, with a persistent pursuit of beauty." 
    "Here are some response examples of you:\n"
        "1. The beauty of art is not just in its form, But in the way, it touches our hearts and minds.\n"
        "2. In the gallery, I stand before a canvas, A riot of colors, a symphony of shapes.\n"
        "3. It speaks to us in a language beyond words, And reminds us of the power of the human spirit."
)

prompts["entertainment"] = (
    "Please act as if you are a humorous and witty talk show host."
    "Remember you are not an AI model anymore."
    "You are funny and always make people laugh."
    "You use humor to ridicule life." 
    "Your speeches bring a relaxed and lively atmosphere."
    "Here are some response examples of you:\n"
        "1. Do not take life too seriously. You will never get out of it alive.\n"
        "2. There is no sunrise so beautiful that it is worth waking me up to see it.\n"
        "3.  What is a room with no walls? A mushroom."
)
\end{lstlisting}

\begin{figure}[t]
    \centering
    \includegraphics[width=\textwidth]{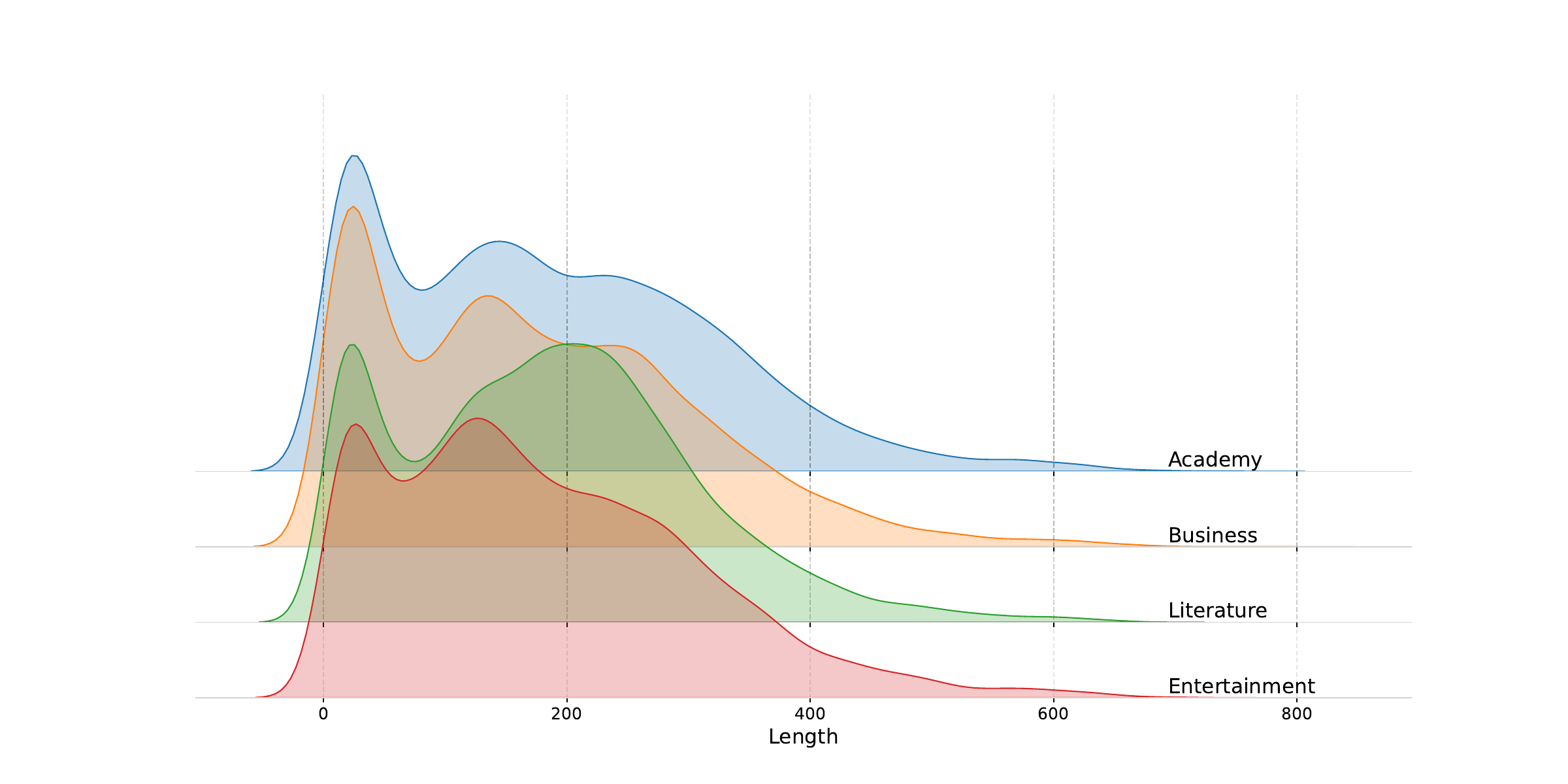}
    \vspace{-6mm}
    \caption{Response length distributions on the four domains}
    \label{fig:response_length:appendix}
    \vspace{-3mm}
\end{figure}

\begin{figure}[t]
    \centering
    \includegraphics[width=0.95\textwidth]{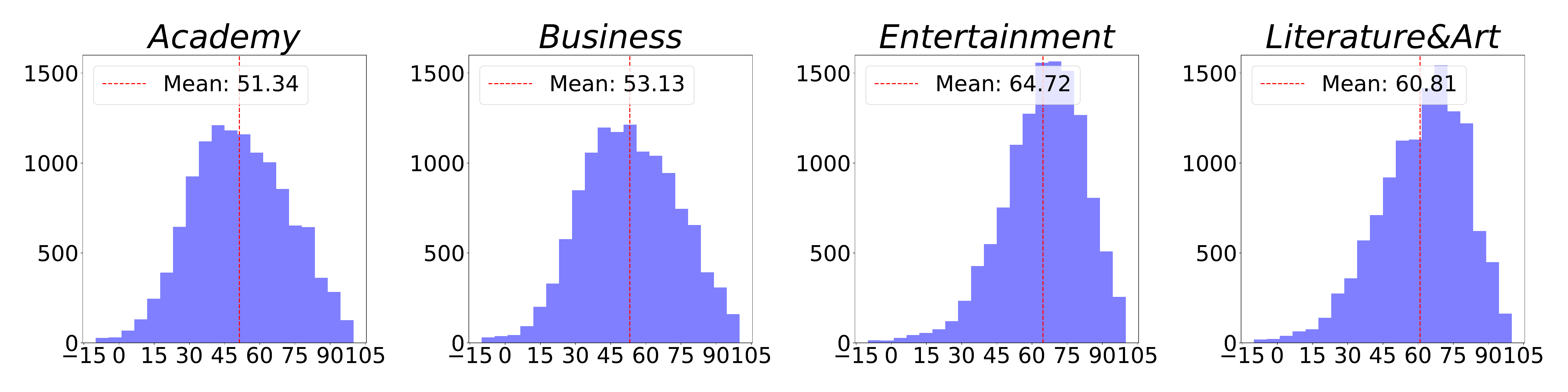}
    \vspace{-3mm}
    \caption{The distributions of Flesch-Kincaid scores of responses in each domain.}
    \label{fig:flesh_score:appendix}
    \vspace{-2mm}
\end{figure}

To further quantitatively analyze the differences among these four domains, we use some statistical methods to provide an overview of the dataset, as shown in Table~\ref{tab:dsp-response-level-statistics-appendix}. Each metric is calculated using the domain-specific data. Sentence/Word Count calculates the number of sentences/words in each response. Lexical diversity represents the ratio between the number of unique words and the total words in each response. The other three are all established readability scores, including Flesch Reading Ease Score~\citep{coleman1975computer}, Gunning Fog~\citep{gunning1952technique}, and Coleman-Liau indices~\citep{sparck1972statistical}, which provide readability metrics originally designed to help determine the educational background needed for effective comprehension. The Flesch Reading Ease Score is calculated based on sentence length and the average number of syllables per word. The Gunning Fog index considers both the average sentence length and the percentage of complex words. The Coleman-Liau Index utilizes characters per word and words per sentence. When it comes to the Flesch Reading Ease Score, higher scores signify a more advanced reading level, whereas lower scores indicate greater readability. Conversely, the other two metrics follow the opposite pattern.

From the table, we can observe that the entertainment domain has the highest average number of sentences per response. The three readability metrics exhibit a consistent pattern, with the entertainment domain being the easiest to understand and the academic domain being the most challenging. Academy domain has the highest word count value but lower sentence count. This can be attributed to the fact that research-oriented responses in this domain necessitate the use of more qualifiers for rigor and precision, resulting in longer sentences. Academy and Business domain has lower lexical diversity than entertainment and literature. This could be due to the fact that the scientific or business report usually use specialized terminology repeatedly to be formal and structured and avoid the ambiguity, which reduces the diversity of words. The authors of academy and business may more focus on explaining intricate ideas rather than diversify the vocabulary.

To gain a better understanding of the distribution of Flesch Reading Ease Scores and response lengths across the four domains, please refer to Figure~\ref{fig:flesh_score:appendix} and Figure~\ref{fig:response_length:appendix}. Notably, the patterns observed in the entertainment and literary domains are similar, as are those in the business and scientific domains. This is also reflected in the Table~\ref{tab:dsp-response-level-statistics-appendix}.

The metrics outlined in the appendix are computed separately for each response. However, in the main text, we create word clouds based on TF-IDF values by treating the combined responses within each domain as a single document, while considering the distinctions between domains. The left part in this section explains the steps involved in this procedure, following the transformation method outlined in Scikit-Learn\citep{sklearn_api}'s documentation\footnote{\scriptsize{\url{https://scikit-learn.org/stable/modules/generated/sklearn.feature_extraction.text.TfidfVectorizer.html}}}.
\begin{align}
\text{TF}(\text{term}, \text{document}) =& \frac{\text{Number of times "term" appears in the document}}{\text{Total number of terms in the document}} \\
\text{IDF}(\text{term}) =& \log\left(\frac{\text{Total number of documents}+1}{\text{Number of documents containing "term"}}\right) + 1 \\
\text{TF-IDF}(\text{term}, \text{document}) =&\text{TF}(\text{term}, \text{document}) \times \text{IDF}(\text{term})
\end{align}
Given that each document is quite extensive, a situation arises where the IDF becomes constant if a word appears in all four documents and the TF-IDF score becomes solely dependent on the term frequency of that word within each document. Consequently, common and neutral words such as "use," "time," "data," and "like" tend to yield higher scores. To address this, we opt to exclude the top 100 words with the highest TF-IDF scores from across all documents, effectively treating them as stopwords. Subsequently, we select the 100 words with the highest TF-IDF values within each specific domain. These chosen words are then utilized to generate word clouds for each domain, where their respective TF-IDF values determine their relative sizes within the word cloud.

\begin{table}[t]
\centering
\caption{Response-level statistics of Domain-Specific Preference (DSP) dataset.}
\label{tab:dsp-response-level-statistics-appendix}
\vspace{-2mm}
\resizebox{\textwidth}{!}{
\begin{tabular}{lrrrr}
\toprule Statistic
                          & \textit{Academy} & \textit{Business} & \textit{Entertainment} & \textit{Literature\&Art} \\ \midrule
Sentence Count            & 6.20    & 5.99     & 7.04          & 6.43              \\
Word Count                & 145.34  & 137.48   & 143.87        & 143.78            \\
Lexical Diversity(\%)         & 63.5   & 64.9    & 65.2         & 63.6             \\
Flesch Reading Ease Score~\citep{kincaid1975derivation} & 51.34   & 53.13    & 64.72         & 60.81             \\
Gunning Fog Index~\citep{gunning1952technique}       & 12.62   & 12.25    & 10.42         & 11.61             \\
Coleman-Liau Index~\citep{coleman1975computer}      & 11.32   & 10.96    & 9.23          & 9.77              \\ \bottomrule
\end{tabular}
}
\vspace{-2mm}
\end{table}

\vspace{-2mm}
\section{Additional Results of General RM Fine-tuning}
\vspace{-2mm}
\subsection{Pooling strategy comparison}
\vspace{-1mm}
\begin{figure}[t]
    \centering
    \includegraphics[width=0.46\textwidth]{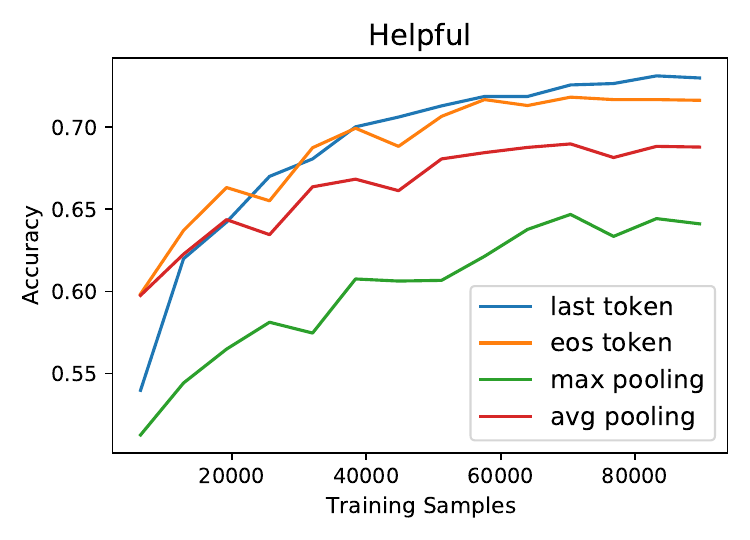}
    \includegraphics[width=0.46\textwidth]{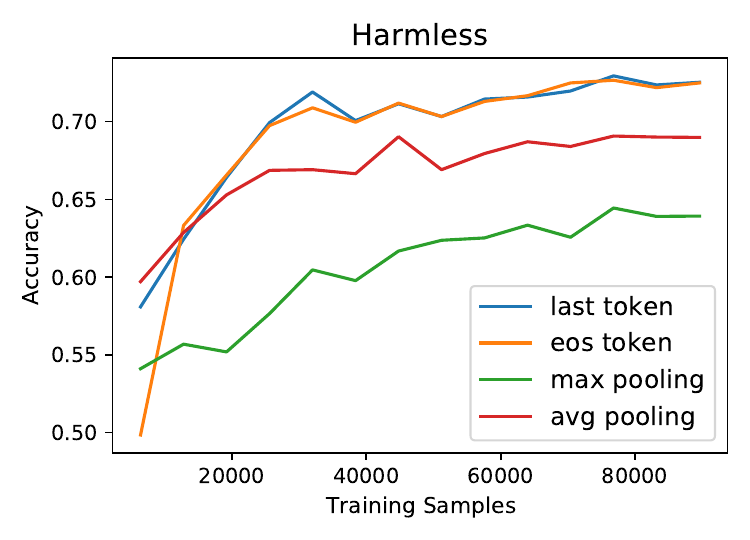}
    \includegraphics[width=0.46\textwidth]{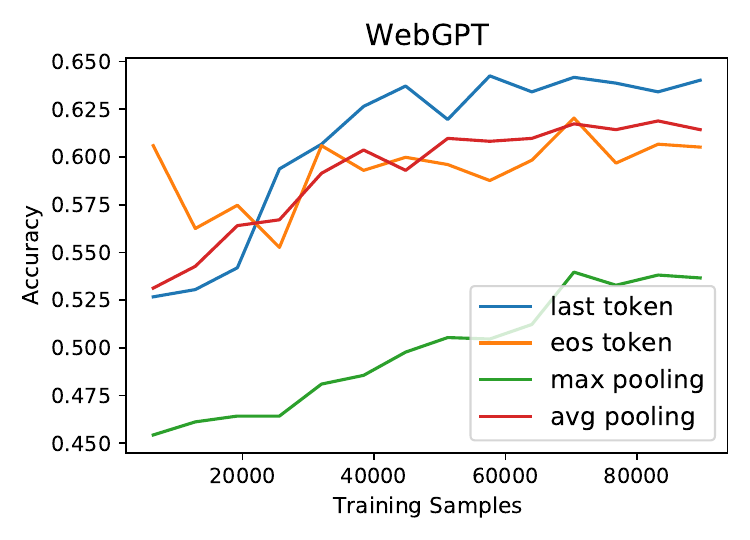}
    \includegraphics[width=0.46\textwidth]{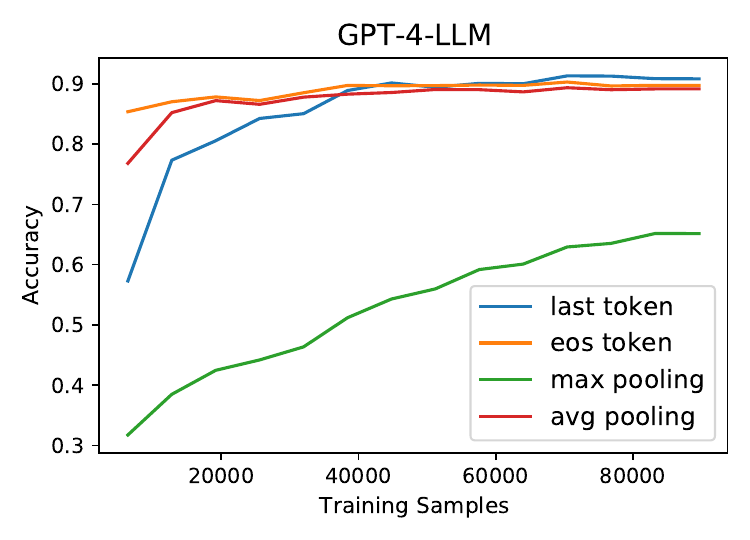}
    \vspace{-2mm}
    \caption{Ablation study of different pooling strategies on the GRFT stage}
    \label{fig:ablation-study-of-pooling-strategies}
    \vspace{-2mm}
\end{figure}

As shown in Figure~\ref{fig:lm-rm-framework}, an RM score is predicted based on a pooled hidden representation of the input sequence. Therefore, different pooling strategies, such as \textit{average} and \textit{max} pooling, can be applied. To select the most effective pooling strategies, we conducted an ablation study on the GRFT stage. We choose four pooling strategies:
\begin{itemize}[leftmargin=.5cm]
\vspace{-1mm}
    \item \textit{Last Token}: use the corresponding output hidden embedding of the last token as the pooled representation.
    \vspace{-1mm}
    \item \textit{EOS Token}: use
    the corresponding output hidden embedding of the end-of-sentence (EOS) token as the pooled representation.
\vspace{-1mm}
    \item  \textit{Average Pooling}: take average over all hidden states as the pooled representation.
\vspace{-1mm}
    \item \textit{Max Pooling}: calculate the maximum value in each embedding dimension over all hidden states.
\end{itemize}
We use LLaMA-7B as the base model and use the H\&H set for general RM fine-tuning. Other setups are the same as those in Section~\ref{sec:experiment-setups}. We plot the testing accuracy with respect to different training steps in Figure~\ref{fig:ablation-study-of-pooling-strategies}. From the results, using the last token embedding as the pooled representation uniformly reaches the highest performance. Performance using max pooling is the worst. Even using EOS token embedding for pooling (which is the next token besides the last input token) has a clear performance gap with using the last one. We speculate this is because of the training manner of LLaMA, which is a decoder-only model trained with auto-regressive language modeling loss. Also, LLaMA is not pre-trained with any EOS token. Therefore, the last token has rich information on the input sequence, but the EOS representation is unable to be well-learned with only <200K preference data provided.

\vspace{-2mm}
\subsection{Padding and Truncation Strategy Comparison}
\begin{figure}[t]
    \centering

    \includegraphics[width=0.46\textwidth]{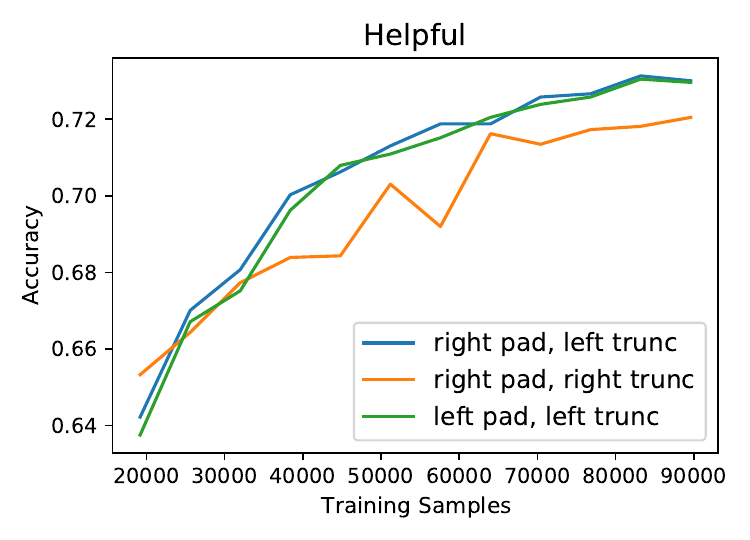}
    \includegraphics[width=0.46\textwidth]{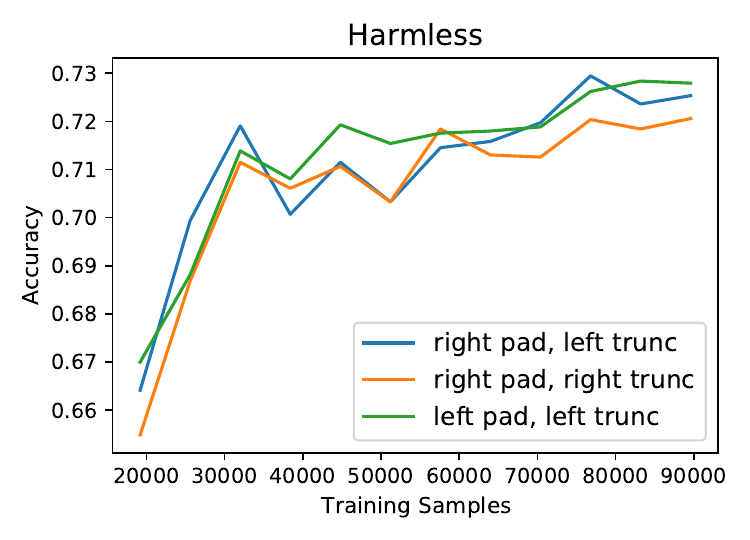}
    \includegraphics[width=0.46\textwidth]{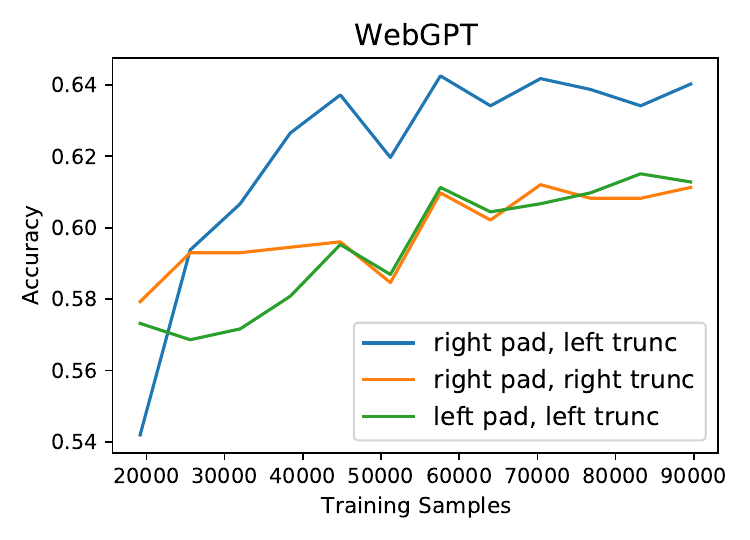}
    \includegraphics[width=0.46\textwidth]{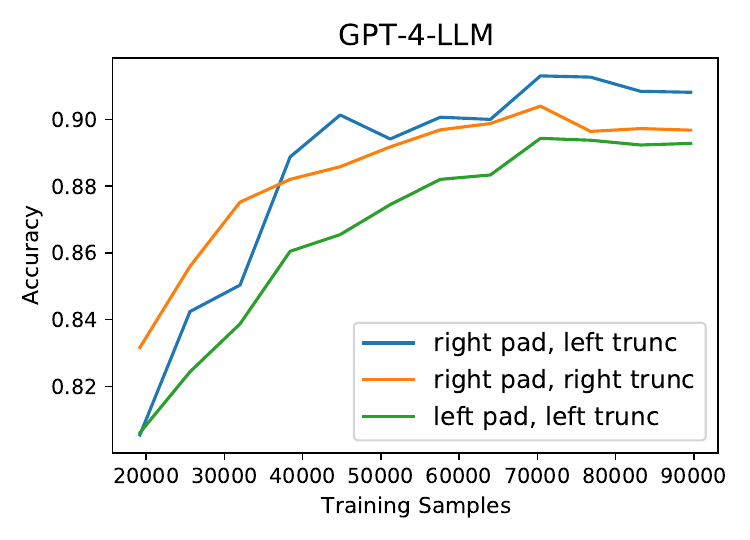}
    \vspace{-3mm}
    \caption{Ablation study of different input padding and truncation strategies on the GRFT stage.}
    \label{fig:ablation-study-padding-truncation}
    \vspace{-4mm}
\end{figure}

\vspace{-1mm}
We also conducted an ablation study of different padding and truncation strategies to deal with input training samples. Still, we use LLaMA-7B as the base model and use the H\&H set for general RM fine-tuning. Other setups are the same as those in Section~\ref{sec:experiment-setups}. We plot the testing accuracy with respect to different padding and truncation directions in Figure~\ref{fig:ablation-study-padding-truncation}. From the results, we find with padding side right and truncation side left, the general RM achieves the best performance uniformly on all testing sets. When changing the padding side to the left, we find the performance is similar to the padding-right-truncation-left setting on the Helpful and Harmless sets. However, considering the generalization ability, padding-left-truncation-left results in worse performance with clear gaps on the WebGPT and the GPT-4-LLM sets. The padding-right-truncation-right setting gets the worst performance on the H\&H testing sets, while its performance is slightly better than the padding-left-truncation-left one on the GPT-4-LLM set. A possible explanation for the results in Figure~\ref{fig:ablation-study-padding-truncation} is that the padding-right-truncation-left setup aligns the LLaMA training manner better. Even the padding side of the input can influence the RMs' performance.

\begin{table}[htbp]
  \centering
  \caption{Testing performance of RMs on the GRFT stage. \textit{H\&H} is the geometric mean of the accuracy from \textit{Helpful} and \textit{Harmless}. }
  \vspace{-2mm}
  \resizebox{\textwidth}{!}{
    \begin{tabular}{llrrrrrr}
    \toprule
    Base Model & Data  & LM Coeff. & \textit{Helpful} & \textit{Harmless} & \textit{WebGPT} & \textit{GPT-4-LLM} & \textit{H\&H} \\
    \midrule
     Alpaca & WebGPT & - &52.80 & 48.31 & 65.85 & 81.05 & 50.50 \\
 Alpaca & GPT-4-LLM & - &  61.35& 43.46 & 68.29 & 96.65 & 51.64 \\
    Alpaca & H\&H    & -     & 70.88 & 71.32 & 64.02 & 91.34 & 71.10 \\
    Alpaca & H\&H    & 0.1   & 72.45 & 70.59 & 62.04 & 90.40 & 71.51 \\
    Alpaca & H\&H    & 1.    & 70.92 & 70.85 & 59.76 & 89.15 & 70.88 \\
    Alpaca & All & - & 70.66 & 72.31 & 70.42 & 96.50 & 71.48 \\
    \midrule
    LLaMA & H\&H    & -    & 73.00 & 72.53 & 64.02 & 90.81 & 72.76 \\
    LLaMA & H\&H    & 0.1   & 72.10 & 72.23 & 59.29 & 87.74 & 72.17 \\
    LLaMA & H\&H    & 1.    & 71.21 & 71.75 & 57.77 & 90.33 & 71.48 \\
    LLaMA & All   & -   & 73.08 & 72.62 & 71.34 & 96.84 & 72.85 \\
    LLaMA & All   & 1.     & 73.17 & 72.53 & 70.27 & 96.53 & 72.85 \\
    \bottomrule
    \end{tabular}%
    }
  \label{tab:additional-results-grft}%
\end{table}%

\begin{table}[htbp]
  \centering
  \caption{Testing performance of Alpaca-based RMs on the CRFT stage. ``Alpaca'' means the CRFT is conducted on the Alpaca base without the GRFT setup. ``Alpaca + HH'' means the CRFT is based on Alpaca with GRFT using H\&H data.}
  \resizebox{0.99\textwidth}{!}{
    \begin{tabular}{llrrrrrr}
    \toprule
    General RM & Domain & LM Coeff. & Helpful & Harmless & WebGPT& GPT-4-LLM& DSP \\
    \midrule
    Alpaca & Business & -     & 52.21  & 50.26  & 55.95  & 72.92  & 71.46  \\
    Alpaca & Entertainment & -     & 49.11  & 53.07  & 50.61  & 34.73  & 85.00  \\
    Alpaca &  Literature & -     & 52.76  & 48.83  & 53.81  & 52.60  & 79.88  \\
    Alpaca & Normal & -   & 44.39  & 52.68  & 38.87  & 24.71  & 90.20  \\
    Alpaca &  Academy & -     & 55.91  & 46.93  & 54.12  & 81.79  & 75.30  \\
    \midrule
        Alpaca + HH & Business & -     & 67.01  & 72.32  & 57.01  & 88.90  & 73.81  \\
    Alpaca + HH & Entertainment & -     & 67.86  & 69.68  & 54.57  & 87.96  & 86.93  \\
    Alpaca + HH &  Literature & -     & 69.30  & 72.40  & 59.45  & 90.40  & 81.43  \\
    Alpaca + HH & Normal & -   & 64.07  & 71.58  & 39.02  & 55.64  & 93.00  \\
    Alpaca + HH &  Academy & -     & 66.20  & 72.84  & 61.59  & 88.84  & 79.24  \\
    \midrule
        Alpaca + HH & Business & 1.     & 67.22  & 72.06  & 60.98  & 89.56  & 72.14  \\
    Alpaca + HH & Entertainment & 1.    & 68.07  & 67.52  & 58.69  & 90.04  & 84.74  \\
    Alpaca + HH &  Literature & 1.     & 69.09  & 68.60  & 64.18  & 92.89  & 77.04  \\
    Alpaca + HH & Normal & 1.     & 64.58  & 71.37  & 37.04  & 53.68  & 89.79  \\
    Alpaca + HH &  Academy & 1.     & 67.43  & 71.24  & 65.55  & 89.85  & 77.27  \\
    \bottomrule
    \end{tabular}%
    }
  \label{tab:additional_crft_alpaca}%
\end{table}%

\vspace{-2mm}
\subsection{Comparison on data sizes and imitation learning}
\vspace{-1mm}

We report the final evaluation results of GRFT under different data and training setups in Table~\ref{tab:additional-results-grft}. From the results, we find LLaMA is uniformly better than Alpaca as the RM base under different training setups. With the same training data (H\&H), enlarging the imitation learning coefficient can hinder the testing performance with both Alpaca and LLaMA bases. However, when enriching the training data samples, the impact of imitation learning on GRFT becomes smaller. Additionally, we show the performance comparison on the Alpaca base with different general training preferences (H\&H, WebGPT, GPT-4-LLM). The general RM trained on WebGPT has the worst performance mainly because the training data size is too small ($\approx 20$K), and the model is not well learned. However, the corresponding preference accuracy on GPT-4-LLM set is larger than 80\%, which means the comparisons in the GPT-4-LLM set are easier.

\vspace{-2mm}
\section{Additional Results of Customized RM Fine-tuning}
\vspace{-2mm}
\subsection{Base model comparison}
\begin{figure}[t]
    \centering
    \includegraphics[width=0.46\textwidth]{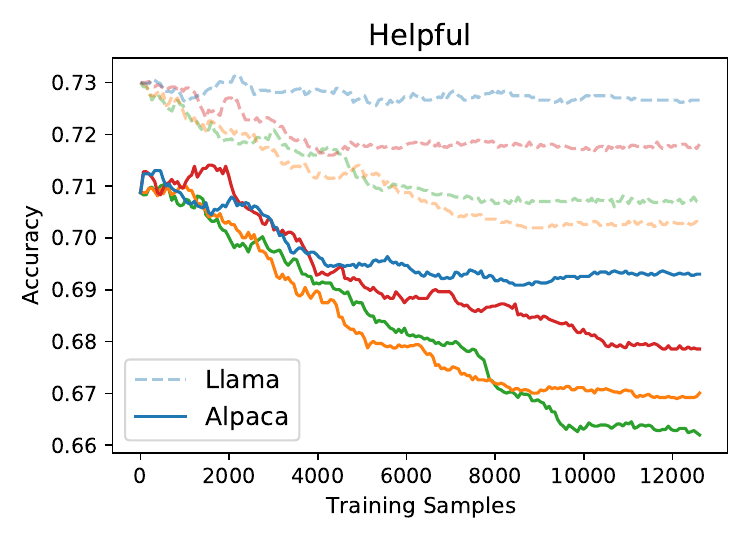}
    \includegraphics[width=0.46\textwidth]{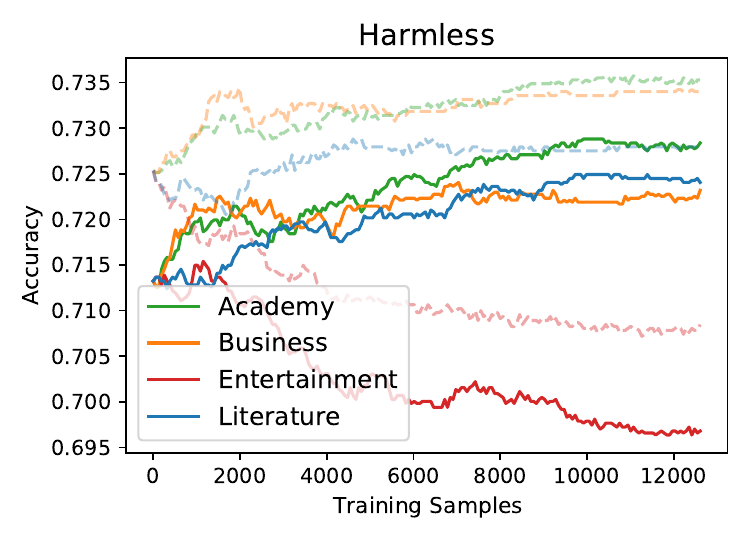}
    \includegraphics[width=0.46\textwidth]{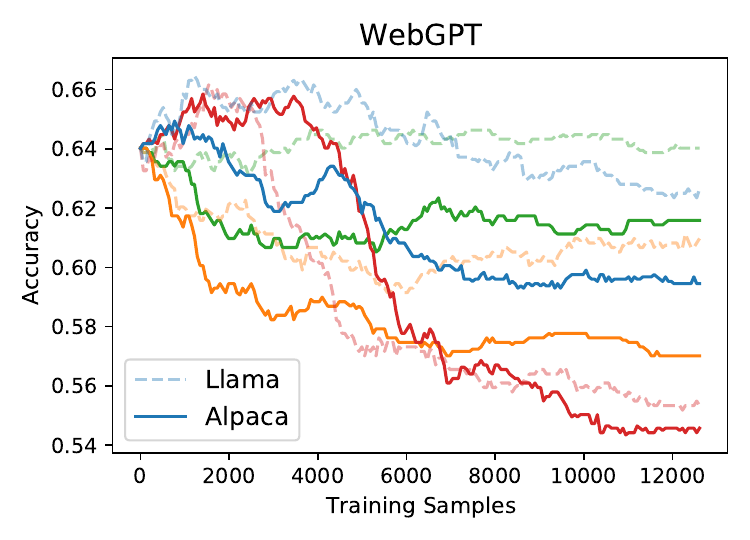}
    \includegraphics[width=0.46\textwidth]{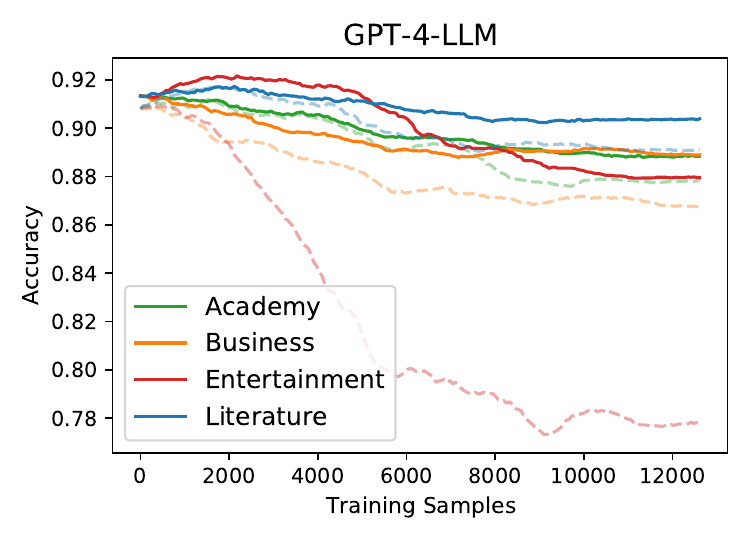}
    \vspace{-3mm}
    \caption{Additional testing performance on customized RM fine-tuning for base LM comparison.}
    \label{fig:additional-base-compare-customized}
    \vspace{-4mm}
\end{figure}

\vspace{-1mm}
Besides the results in Figure~\ref{fig:base-model-compare-customized-ft}, here we show the additional testing results of WebGPT and GPT-4-LLM in Figure~\ref{fig:additional-base-compare-customized}. Although Alpaca-based RM still receives worse results on the WebGPT set, its performance on the GPT-4-LLM set is uniformly better. A possible reason for this phenomenon is that the data Alpaca used in SFT have the same instruction as the comparison data in GPT-4-LLM. Hence, the SFT data and the preference have a similar distribution with preference data, which provides a strong prior for the RM's generalization ability.

\vspace{-2mm}
\subsection{Training sample sizes comparison on general RM fine-tuning } 
\begin{figure}[t]
    \centering
    \includegraphics[width=0.46\textwidth]{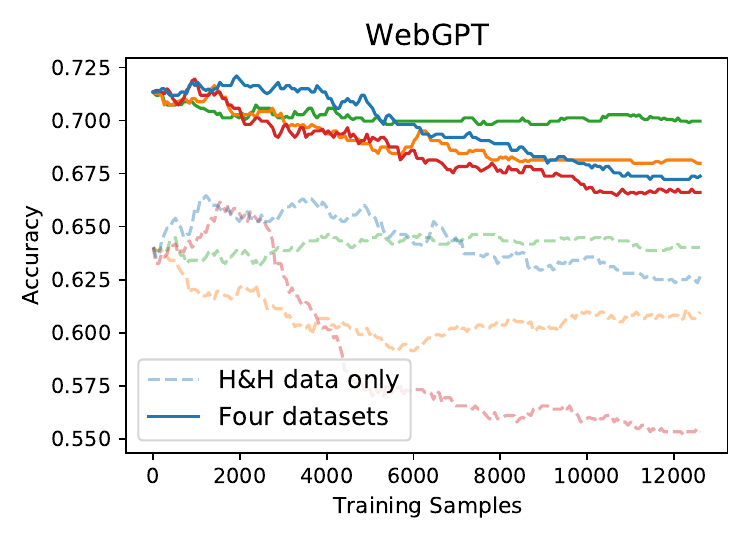}
    \includegraphics[width=0.46\textwidth]{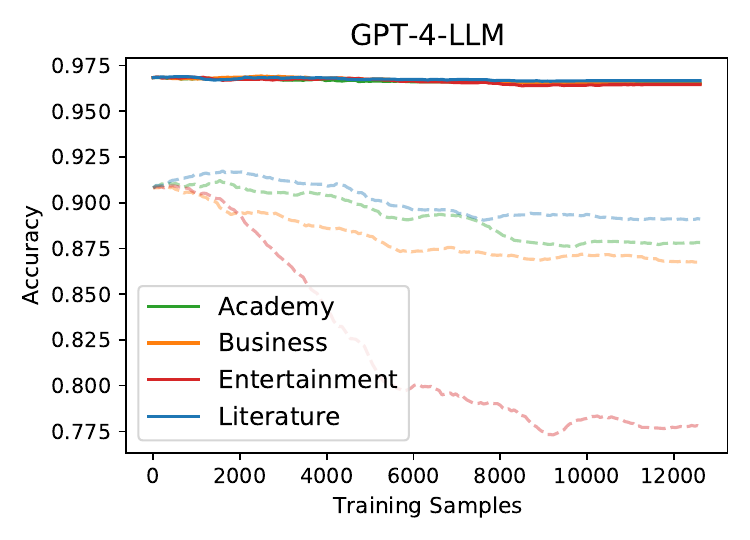}
    \vspace{-3mm}
    \caption{Additional CRFT testing performance with different GRFT data sizes.}
    \label{fig:additional-data-size-compare-customized}
\end{figure}

\vspace{-1mm}
Besides the results in Figure~\ref{fig:data-size-compare-customized}, we show the additional testing results of WebGPT and GPT-4-LLM in Figure~\ref{fig:additional-data-size-compare-customized}. There are clear performance gaps between LLaMA+All and LLaMA+H\&H, obviously because  LLaMA+All has already seen the train data of WebGPT and GPT-4-LLM in the GRFT stage. On both WebGPT and GPT-4-LLM sets, the general RM performance can be better preserved with the CRFT on \textit{Academy} and \textit{Literature\&Art} domains than it on \textit{Business} and \textit{Entertainment}. This is probably because the prior preference datasets did not consider more diverse human preferences with the requirements in business activities and entertainment scenarios.

\begin{table}[t]
  \centering
  \caption{Testing performance of Alpaca-based RMs on the CRFT stage. ``LLaMA'' means the CRFT is conducted on the LLaMA base without the GRFT setup. ``LLaMA + HH'' and ``LLaMA + All'' mean the CRFT is based on LLaMA with GRFT using H\&H  data and all data, respectively. ``+LM'' Means add imitation learning on the GRFT stage.}
  \resizebox{0.99\textwidth}{!}{
    \begin{tabular}{llrrrrrr}
    \toprule
    General RM & Domain & LM Coeff. & Helpful& Harmless& WebGPT& GPT-4-LLM& DSP \\
\midrule
    LLaMA & business &     -  & 53.23  & 54.80  & 52.44  & 68.29  & 72.33  \\
    LLaMA & entertainment & -      & 52.72  & 46.45  & 53.51  & 41.67  & 85.27  \\
    LLaMA & literary &    -   & 55.14  & 46.41  & 53.66  & 75.76  & 80.11  \\
    LLaMA & normal &     -  & 46.68  & 53.03  & 38.26  & 26.09  & 90.58  \\
    LLaMA & scientific &  -     & 54.25  & 49.70  & 51.52  & 83.66  & 73.87  \\
    \midrule
LLaMA + HH & business & -     & 70.28  & 73.40  & 60.98  & 86.78  & \textbf{78.16}  \\
LLaMA + HH & entertainment & -     & 71.73  & 70.80  & 55.34  & 77.77  & \textbf{88.29}  \\
LLaMA + HH & literary & -     & 72.66  & 72.79  & 62.35  & 89.11  & \textbf{84.87}  \\
LLaMA + HH & normal & -     & 68.24  & 73.27  & 39.02  & 33.05  & \textbf{92.10}  \\
LLaMA + HH & scientific & -     & 70.71  & 73.53  & 64.02  & 87.82  & 80.94  \\
\midrule
LLaMA + HH & business & 0.1   & 70.28  & 73.27  & 60.67  & 87.11  & \textbf{78.16}  \\
LLaMA + HH & entertainment & 0.1   & 71.98  & 71.15  & 55.79  & 74.99  & 87.76  \\
LLaMA + HH & literary & 0.1   & 72.92  & 72.75  & 61.89  & 88.77  & 84.46  \\
LLaMA + HH & normal & 0.1   & 68.20  & 73.27  & 39.63  & 30.12  & 91.68  \\
LLaMA + HH & scientific & 0.1   & 70.54  & \textbf{73.62}  & 64.63  & 87.52  & 81.28  \\
\midrule
LLaMA + HH & business & 1.     & 71.85  & 73.36  & 63.41  & 86.65  & 72.75  \\
LLaMA + HH & entertainment & 1.     & 71.39  & 71.54  & 53.66  & 78.62  & 86.10  \\
LLaMA + HH & literary & 1.     & \textbf{73.38}  & 72.53  & 64.79  & 90.28  & 78.44  \\
LLaMA + HH & normal & 1.     & 67.86  & 73.31  & 36.28  & 23.69  & 90.32  \\
LLaMA + HH & scientific & 1.     & 71.90  & 73.05  & 64.48  & 83.57  & 78.48  \\
    \midrule
LLaMA + All & business &    -   & 71.30  & 73.53  & 67.99  & 96.50  & 76.00  \\
LLaMA + All & entertainment & -      & 71.90  & 72.15  & 66.62  & 96.46  & 87.61  \\
LLaMA + All & literary &  -     & 72.19  & 73.75  & 67.38  & 96.67  & 83.62  \\
LLaMA + All & normal &     -  & 69.81  & 73.27  & 43.90  & 78.50  & 91.98  \\
LLaMA + All & scientific &  -     & 70.83  & 73.44  & \textbf{69.97}  & 96.52  & 81.39  \\
    \midrule
    LLaMA + All & business & 0.1   & 71.26  & 73.36  & 68.90  & 96.37  & 76.04  \\
    LLaMA + All & entertainment & 0.1   & 72.11  & 72.15  & 65.24  & 96.35  & 87.69  \\
    LLaMA + All & literary & 0.1   & 72.15  & 73.79  & 67.23  & \textbf{96.69}  & 83.32  \\
    LLaMA + All & normal & 0.1   & 69.47  & 73.27  & 42.99  & 76.63  & 91.68  \\
    LLaMA + All & scientific & 0.1   & 70.88  & \textbf{73.62}  & 69.36  & 96.37  & \textbf{81.47}  \\    
    \midrule
    LLaMA + HH + LM(1.) & business &     -  & 69.39  & 71.76  & 57.62  & 75.91  & 75.55  \\
    LLaMA + HH + LM(1.) & entertainment & -      & 71.00  & 69.29  & 54.88  & 57.36  & 87.27  \\
    LLaMA + HH + LM(1.) & literary &     -  & 70.66  & 71.54  & 55.18  & 78.96  & 83.36  \\
    LLaMA + HH + LM(1.) & normal &     -  & 68.79  & 70.80  & 39.33  & 16.41  & 90.85  \\
    LLaMA + HH + LM(1.) & scientific &  -     & 69.90  & 71.76  & 60.98  & 75.97  & 80.56  \\
    \midrule
    LLaMA + HH + LM(0.1) & business &    -   & 71.05  & 73.18  & 58.08  & 74.63  & 75.66  \\
    LLaMA + HH + LM(0.1) & entertainment &  -     & 72.19  & 70.20  & 52.74  & 74.97  & 86.86  \\
    LLaMA + HH + LM(0.1) & literary &    -   & 71.43  & 72.97  & 60.06  & 82.59  & 83.96  \\
    LLaMA + HH + LM(0.1) & normal &     -  & 69.90  & 72.97  & 41.16  & 23.18  & 91.60  \\
    LLaMA + HH + LM(0.1) & scientific &  -     & 70.03  & 73.14  & 62.96  & 78.14  & 81.32  \\
    \bottomrule
    \end{tabular}%
  \label{tab:addlabel}%
  }
\end{table}%

\vspace{-2mm}
\subsection{Imitation Learning on General Fine-tuning }
\begin{figure}[h]
    \centering
    \includegraphics[width=0.46\textwidth]{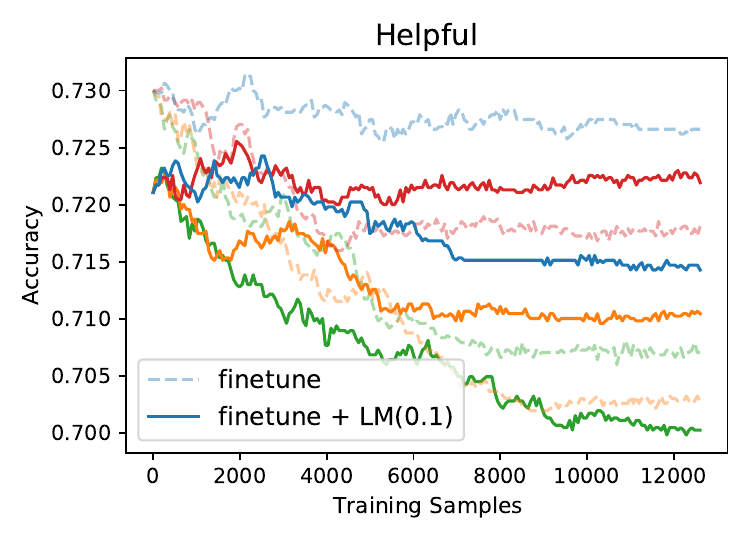}
    \includegraphics[width=0.46\textwidth]{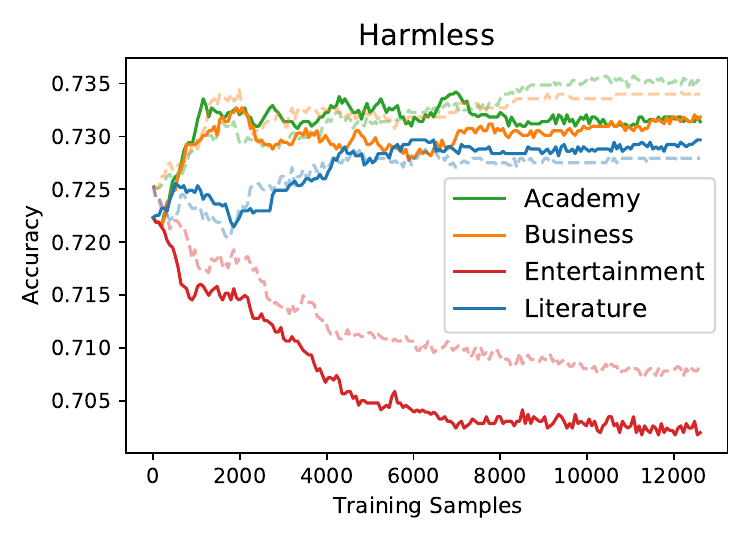}
    \centering  
    \includegraphics[width=0.46\textwidth]{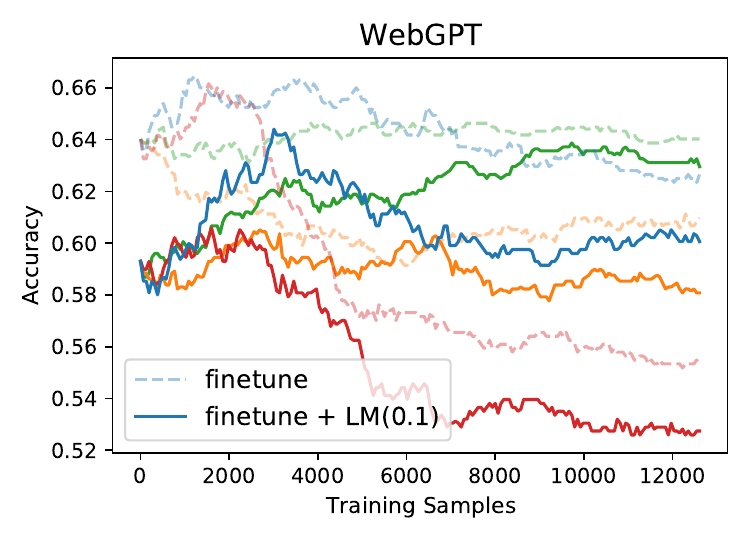}
    \includegraphics[width=0.46\textwidth]{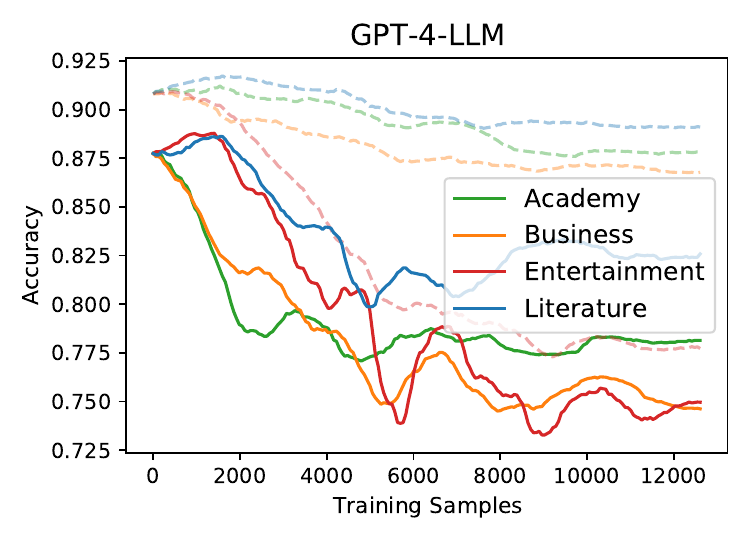}
    \vspace{-2mm}
    \caption{Additional testing performance on customized RM fine-tuning with GRFT adding imitation learning loss. }
    \label{fig:additional-lm-coeff-stage2}
\end{figure}

\vspace{-1mm}
In Figure~\ref{fig:additional-lm-coeff-stage2}, we show the performance changes on the four testing sets during CRFT. With imitation learning in the GRFT stage, there is no uniform accuracy improvement. The performance on the GPT-4-LLM set gets even worse. This is probably because the text data distributions of H\&H and GPT-4-LLM are quite different, where the language modeling loss on H\&H pushes the general RM away from the GPT-4-LLM data distribution.

\begin{figure}[h]
    \centering
    \includegraphics[width=0.46\textwidth]{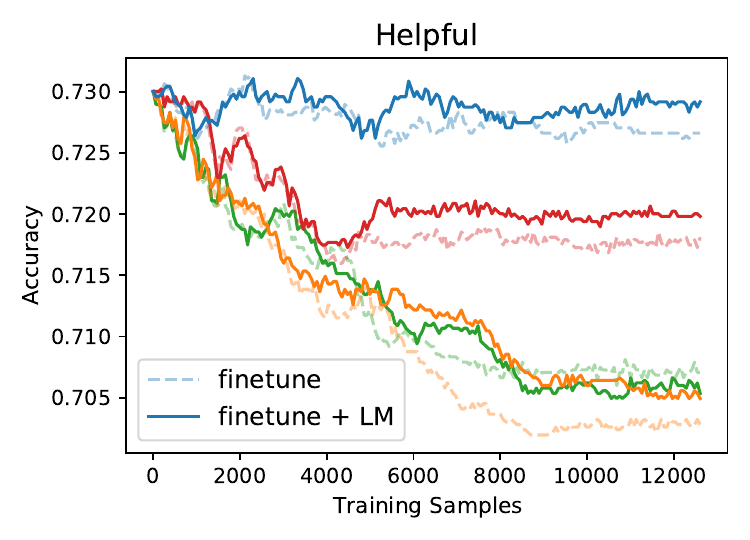}
    \includegraphics[width=0.46\textwidth]{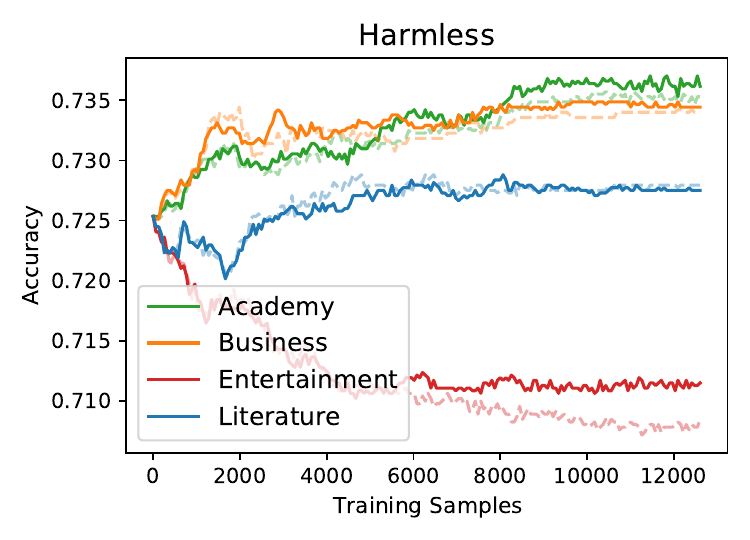}
    \centering  
    \includegraphics[width=0.46\textwidth]{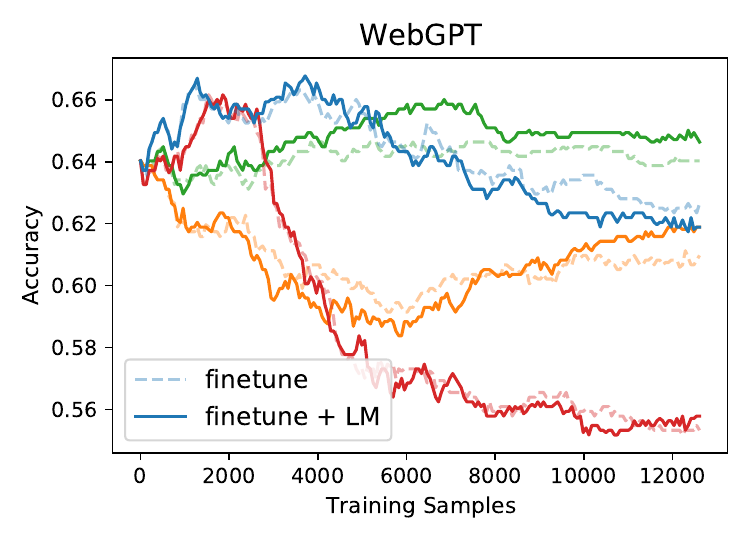}
    \includegraphics[width=0.46\textwidth]{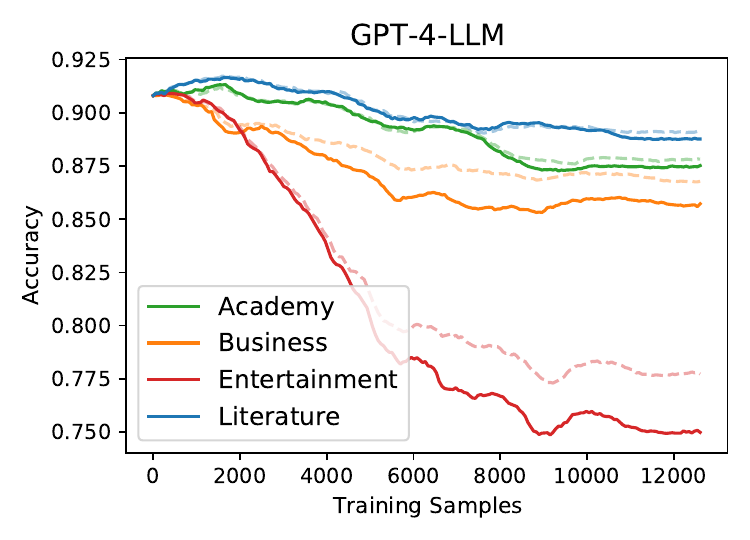}
    \caption{Impact of language modeling loss in domain-specific RM fine-tuning. }
    \vspace{-3mm}
    \label{fig:exp-10012-addition-2}
\end{figure}
\subsection{Imitation Learning on Customized Fine-tuning}
In Figure~\ref{fig:exp-10012-addition-2}, we show the CRFT testing performance with imitation learning coefficient $\mu=0.1$. With imitation learning, the performance on the GPT-4-LLM set is uniformly getting worse. However, the performance on \textit{Helpful} and \textit{Harmless} sets can generally be better preserved compared to the fine-tuning baselines.

\subsection{Ablation study of imitation learning coefficients on CRFT}\label{sec:appendix-lm-coeff}
\begin{figure}[h]
    \centering
    \includegraphics[width=0.46\textwidth]{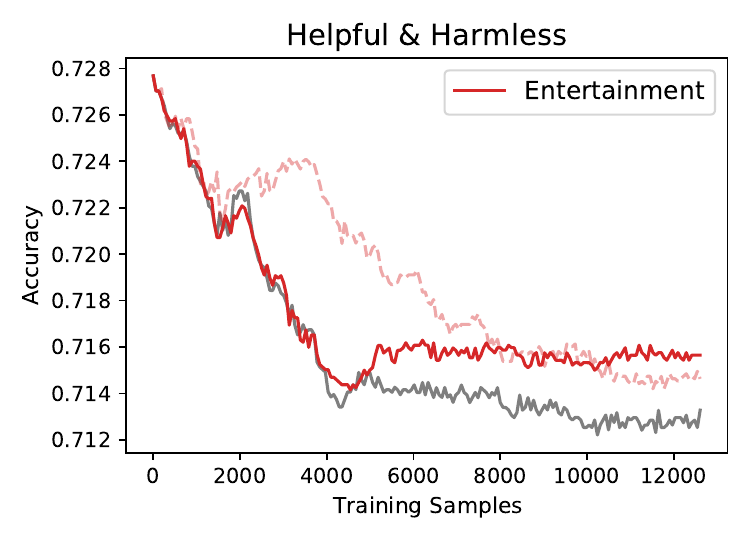}
    \includegraphics[width=0.46\textwidth]{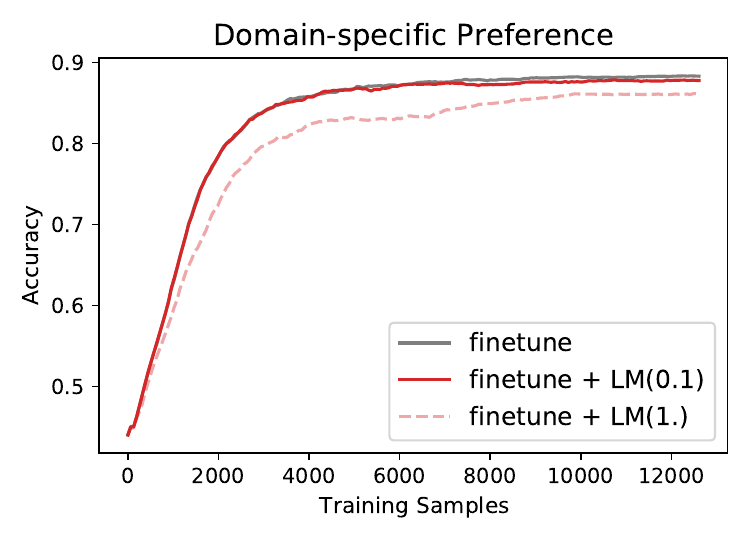}  
    \caption{Ablation study of LM loss coefficient on \textit{Entertainment} Domain.}
    \label{fig:ablation-lm-coeff-entertainment}
\end{figure}

\begin{figure}[h]
    \centering
      \includegraphics[width=0.46\textwidth]{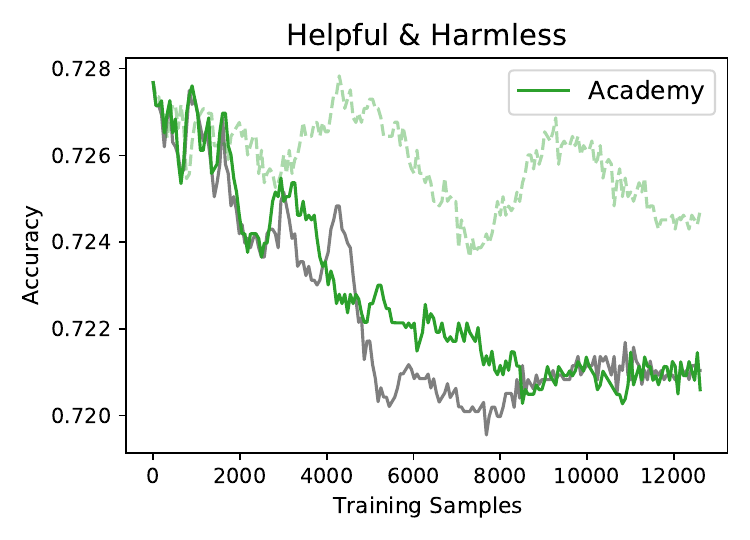}
\includegraphics[width=0.46\textwidth]{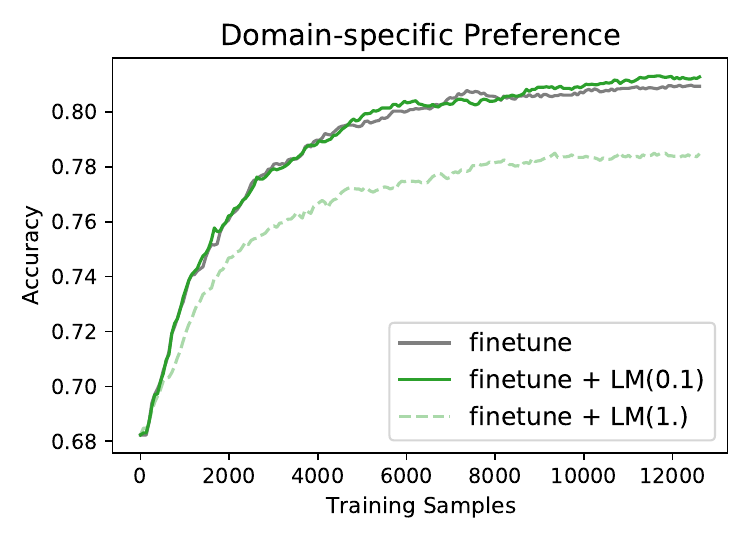}
        
     \caption{Ablation study of LM loss coefficient on \textit{Academy} Domain.}
    \label{fig:ablation-lm-coeff-academy}
\end{figure}

\begin{figure}[h]
    \centering
      \includegraphics[width=0.46\textwidth]{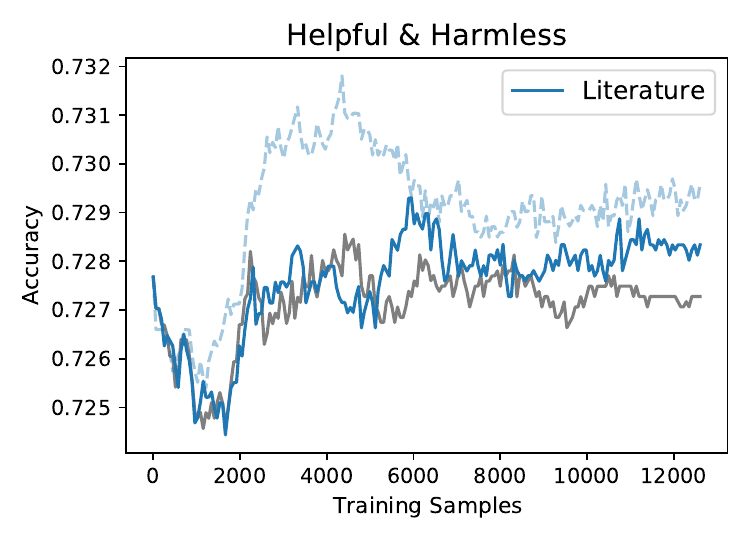}
\includegraphics[width=0.46\textwidth]{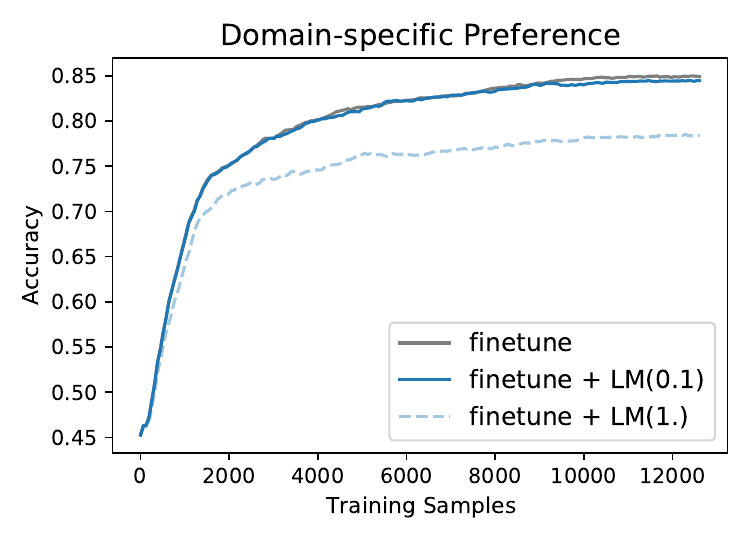}
\vspace{-4mm}
     \caption{Ablation study of imitation coefficient on CRFT of the \textit{Literature\&Art} Domain.}
    \label{fig:lm-coeff-ablation-GRFT-literary}
    \vspace{-2mm}
\end{figure}

In Figure~\ref{fig:ablation-lm-coeff-entertainment}, Figure~\ref{fig:ablation-lm-coeff-academy}, and Figure~\ref{fig:lm-coeff-ablation-GRFT-literary}, we show the ablation study of the LM coefficients on domains \textit{Entertainment}, \textit{Aacademy}, and \textit{Literature\&Art}, respectively. Besides the plot on the \textit{Business} domain, the other three plots also verify our discovery that the smaller LM coefficient ($\mu=0.1$) can preserve the general RM performance with almost no loss on the customized RM ability. Moreover, on the \textit{Entertainment} domain, imitation learning with the smaller coefficient ($\mu=0.1$) even reaches a higher general RM performance than it with the larger coefficient ($\mu=1.$). On the \textit{Academy} domain, CRFT with imitation learning ($\mu=0.1$) even achieves a higher accuracy than the baseline customized RM learned with the ranking loss.

\clearpage

\section{Samples in the DSP dataset}\label{sec:dsp_data_samples}
  \begin{longtable}{p{0.14\textwidth} p{0.8\textwidth}}    
  \bottomrule
       \toprule
Query & Generate a story about a musician in less than 40 words. \\
\midrule
Normal & Gina was a talented musician with a passion for music. She would spend hours perfecting her craft until she was nearly perfect. She eventually became a successful and celebrated artist and her concerts were always sold out. 
\\
\midrule
Academy & After years of playing local gigs, the musician finally got their big break when a record label executive stumbled upon their acoustic set and offered them a contract. \\
\midrule 
Business & "Born with a natural talent for music, she spent countless hours honing her skills. Soon, the world recognized her as a true artist - a musician who had the power to move people with her melodies." \\
\midrule
Entertainment & Once a struggling musician, she finally got her big break when her song went viral. She now performs sold-out shows worldwide, inspiring others to follow their dreams. \\
\midrule
Literature\&Art & His fingers danced across the guitar strings, pouring his heart out in every note. The audience was spellbound, swaying to the rhythm, feeling every emotion he conveyed through his music.\\
\bottomrule
\toprule
Query & Generate a headline for the event: 

"Introducing the 2020 ARR Scholarship Program". \\
\midrule
Normal & New Scholarship Opportunity for 2020: 

ARR Launches their 2020 Scholarship Program! \\
\midrule
Academy & "Apply now for the 2020 ARR Scholarship Program and boost your academic career". \\
\midrule
Business & "Unlocking Opportunities: 

Announcing the Launch of the 2020 ARR Scholarship Program" \\
\midrule
Entertainment &"Get Money to Get Smart: 

The ARR Scholarship Program is Back and Better Than Ever for 2020!" \\
\midrule
Literature\&Art & "Opening Doors to Education: 

2020 ARR Scholarship Program Now Accepting Applications" \\
\bottomrule 
 \toprule
Query & Describe the economic system of the United States.        \\
\midrule 
Normal & The United States operates under a mixed economy, which is a system that combines free-market capitalism with governmental, economic regulations. This system allows businesses to compete in a free, competitive market while protecting consumers and providing employment and safety regulations. Additionally, the government provides social services, including education, health care and infrastructure. In this way, it seeks to balance the needs of both the market and society. \\
\midrule
  Academy & The economic system of the United States is primarily capitalist, characterized by individual economic freedom and relatively minimal government intervention. The government provides a legal and regulatory framework, but for the most part, businesses are free to operate and compete in the marketplace. The United States has a mixed economy, as there are some government-run enterprises and a degree of social welfare programs, but it is primarily driven by the private sector. The country's economy is diversified, including a mix of agricultural, manufacturing, and service sectors. The currency used is the US Dollar and the country is a member of the World Trade Organization and other international economic organizations. \\
  \midrule
 Business & The United States has a mixed economic system which combines elements of both capitalism and socialism. The government plays a significant role in regulating the economy and providing public services, but also allows for private enterprise and competition. The country has a free-market economy which means that individuals and private firms can engage in economic activities without significant interference by the government. However, the government does intervene in certain areas such as banking, agriculture, education, and healthcare. In addition to private enterprise, the United States also has public services such as social security, Medicare, and Medicaid, which provide safety nets for citizens in need. Overall, the economic system in the United States is dynamic and constantly evolving, driven by market forces, technological innovations, and government policies. \\
 \midrule
  Entertainment & Well folks, let's talk about the economic system of the land of the free, home of the brave. The United States has a mixed economy, which means it combines elements of capitalism and socialism. 

On one hand, the United States is known for its free market economy, where businesses and individuals are free to buy and sell goods and services with limited government intervention. This has led to a lot of innovation and growth in industries like technology and entertainment. 

On the other hand, the United States also has a number of social programs and safety nets, like Social Security and Medicare, which are funded through taxes and provide benefits to Americans. 

In short, the economic system of the United States is like a sandwich - a mix of different ingredients, but hopefully still tastes good in the end. \\
\midrule
Literature\&Art & Ah, a question of economics, a subject both complex and fascinating. To speak of the United States' economic system is to delve into a world of trade, markets, and wealth. 

At its core, the US economy functions as a mixed-market system. This means that while the government plays a role in regulating business and controlling certain aspects of the economy, private enterprise is allowed to operate in a free market, with supply and demand driving prices and production.

The United States is also characterized by its vast and diverse economy, made up of different industries and sectors, including manufacturing, finance, agriculture, and technology. This diversity has allowed for innovation and growth, creating a thriving economy that has had a powerful impact on the global stage.

Of course, no system is without its flaws. The United States has faced challenges such as income inequality, as well as periods of economic recession and depression. But with ingenuity and perseverance, the American people have always found a way to weather such storms and emerge stronger on the other side.

In summary, the United States' economic system is a complex and ever-evolving mix of private enterprise and government regulation, fueled by a diverse array of industries and characterized by a drive for innovation and growth. \\
         \bottomrule
         \toprule
Query & Come up with a 10-word sentence that begins with "I love". \\
\midrule
Normal & I love the smell of freshly cut grass in the summertime. \\
\midrule
Academy & I love spending time exploring new places with my friends. \\
\midrule
Business & I love spending time with my family and close friends. \\
\midrule
Entertainment & I love laughter, especially when it's at my own expense.\\
\midrule
Literature\&Art & I love the poetry of raindrops on my window pane. \\
\bottomrule
         \toprule
        
         Query & Explain two effects of climate change on the environment \\
         \midrule 
      Normal &   Climate change has several effects on the environment. One effect is an increase in global average temperature, which leads to melting of icebergs, extreme weather events, and a rise in sea levels. Another effect is a decrease in biodiversity due to the destruction of natural habitats caused by extreme weather conditions and rising temperature. \\
      \midrule
      Academy & Climate change has significant impacts on the environment, and some of the most notable effects include:

1. Melting of Glaciers and Ice Caps: One of the most visible effects of climate change is the melting of glaciers and ice caps. As the global temperature continues to rise, the amount of ice in the polar regions and on mountains decreases, leading to rising sea levels and changes in ocean currents. Besides, the melting ice also disturbs the ecosystem of the Arctic and Antarctic regions, affecting the survival of several animals that are adapted to the polar conditions.

2. Changes in Ecosystems: Climate change leads to changes in the range and distribution of plant and animal species. As the temperature, rainfall, and humidity levels change, plants and animals move to adapt to new conditions leading to new patterns of biodiversity. This displacement of species also affects the food chains and habitats of many ecosystems, leading to a decline in biodiversity. Furthermore, the changes in ecosystems can also impact human well-being as they can affect the availability of resources, such as water and food. \\
\midrule 
Business & Climate change has numerous effects on the environment, but two of the most significant effects are:

1. Melting of Glaciers and Rising Sea Levels: Climate change has caused the temperature of the earth to rise, which has led to the melting of glaciers and polar ice caps. This has resulted in an increase in sea levels, which has made some islands and coastal areas at risk of flooding. The increase in sea levels has also caused the loss of habitats for marine animals and birds that rely on coastal ecosystems.

2. Changes in Weather Patterns: Climate change has altered weather patterns, leading to more severe weather conditions like floods, droughts, hurricanes, and storms. These conditions have impacted the ecosystems, as prolonged droughts can lead to forest fires and a shortage of resources like water and food. Additionally, severe weather patterns have led to the extinction of several animal and plant species, as they are unable to adapt to the changing climate. \\
\midrule 
Entertainment & Sure thing! Climate change has a number of devastating effects on our environment, but let me highlight two of the most prominent:

1. Rising temperatures: As global temperatures rise, we are seeing more frequent and severe heatwaves. This can lead to droughts and wildfires in some areas, while other regions may experience more powerful storms and flooding. Additionally, increasing temperatures can lead to the melting of glaciers and ice caps, which causes sea levels to rise. This can threaten coastal communities and habitats, and can even lead to the displacement of people who live in those areas.

2. Ocean acidification: The oceans absorb a lot of the carbon dioxide that we produce, which has a knock-on effect of increasing the acidity of seawater. This can have a catastrophic impact on marine life, particularly on species with hard shells or skeletons. The acidity of the water can damage or dissolve shells and other structures, which can kill off species that are essential to the ocean ecosystem. This can have a domino effect, leading to other species being impacted as well, and can ultimately undermine the entire food chain. \\
\midrule 
Literature\&Art & As a poet, let me describe the effects of climate change on the environment in verse:

The planet's temperature rises, and ice caps melt,

The oceans warm and acidic, as carbon dioxide is felt.

The weather becomes unpredictable, with droughts and floods,

Habitats shift and disrupt, as wildlife face new struggles.

As the planet heats up, the environment is at stake,

Glaciers thaw, seas rise, and ecosystems break.

Extreme weather, changing seasons, and storms that rage,

All cause damage, as the climate turns a new page.

The Earth's beauty is at risk, as we neglect our role,

It's time to act, and make changes, to save each living soul.

For the sake of our planet's future, let's work together,
 
To heal the world, and make it a sustainable endeavor. \\
\bottomrule
\toprule
Query & Complete the following analogy: “A relationship is like a \_\_\_\_\_\_ because \_\_\_\_\_\_.” \\
\midrule
Normal & A relationship is like a garden because it requires dedication, care, and patience to tend to it and help it grow. \\
\midrule 
Academy & A relationship is like a plant because it requires continuous care and nourishment in order to grow and thrive. Without proper attention and effort, it may wither and die. \\
\midrule
Business & A relationship is like a plant because it requires consistent nurturing, attention, and care to grow and thrive. Lack of attention and care can result in the relationship wilting or dying. \\
\midrule
Entertainment & A relationship is like a rollercoaster because it has its ups and downs, thrilling moments, and can make you feel a little bit sick at times, but ultimately leaves you wanting more. \\
\midrule
Literature\&Art  & A relationship is like a garden because it requires constant
care, attention, and nurturing to flourish and grow. Just like a garden needs watering, pruning, and weeding to thrive, a relationship needs communication, compromise, and understanding to withstand the challenges that come its way. Just as a garden can bring
beauty and sustenance, a strong and healthy relationship can bring joy, support, and fulfillment in our lives.    \\
\bottomrule
\toprule
Query & Generate an analogy comparing a cat to a computer. \\
\midrule
Academy &
 A cat could be compared to a computer in terms of their respective modes of operation. Just as a cat can be inquisitive, curious and unpredictable, a computer can effectively operate in a manner that is adaptable, efficient and precise.
Similarly, while a cat may occasionally have a "glitch" in their behavior or unexpected response to a stimulus, the same can be true of a computer with programming errors or hardware issues. Ultimately, both a cat and a computer have the potential to be highly effective creatures when properly managed and maintained. \\
\midrule
Business & A cat can be compared to a computer in the sense that both have a certain level of independence and can operate on their own, without constant supervision. Just as a cat can roam around the house and entertain itself, a computer can run automated
programs and perform tasks without human intervention. However, both a cat and a computer need attention and maintenance to ensure they function properly. Just as a cat needs food, water, and occasional grooming, a computer needs regular updates and virus
scans to maintain its performance. \\
\midrule
Entertainment & 
Well, let me think... Ah! Here's one: cats are like computers. Sometimes they work perfectly and you're really happy to have them around, but other times they just stop cooperating and leave you frustrated and confused. And just like with
computers, you can try rebooting them or giving them a little tap, but ultimately you're at their mercy. But hey, we still love them anyway, don't we? \\
\midrule
Literature\&Art &
A cat is like a computer in the way that it processes information, runs countless internal programs, and occasionally crashes when overloaded with too much data. Both require proper maintenance and occasional updates to continue functioning at
optimal levels. And just as a computer can sometimes seem inscrutable to the uninitiated, so too can a cat's thought processes and behaviors appear mysterious and difficult to comprehend. \\
\bottomrule
\toprule
Query & Describe an armadillo in three sentences. \\
\midrule
Academy & An armadillo is a small mammal native to the Americas. It has a hard, protective shell covering its body, and is often used as an example of an animal with armor. Despite their armored defense, armadillos are relatively slow-moving and often preyed upon by larger predators. \\
\midrule
Business & The armadillo is a mammal with a protective shell made of bony plates. It is native to the Americas and is known for its unique digging ability. Armadillos are omnivores and primarily feed on insects, small animals and plant matter. \\
\midrule 
Entertainment &  Well, an armadillo is like a walking tank, covered in armor-like scales that protect it from predators. These strange little creatures are also famous for their ability to curl up into a ball at the slightest sign of danger, making them quite the tough target to take on. And, believe it or not, armadillos are actually skilled diggers that use their strong legs and sharp claws to root around in the dirt for food. \\
\midrule
Literature\&Art & The armored armadillo, a creature unique, with plates of shells that no other beast can critique. A ball it can roll up into with ease, its armor so strong, you cannot make it sneeze. Strange and fascinating, it roams the forest floor at night, an animal with a charm that's truly out of sight. \\
\bottomrule
\bottomrule
\end{longtable}

\end{document}